\documentclass{article}

\PassOptionsToPackage{numbers, compress}{natbib}
\usepackage[preprint]{neurips_2026}
\usepackage[utf8]{inputenc}
\usepackage[T1]{fontenc}
\usepackage{hyperref}
\usepackage{url}
\usepackage{booktabs}
\usepackage{amsfonts}
\usepackage{nicefrac}
\usepackage{microtype}
\usepackage{xcolor}
\usepackage{amsmath}
\usepackage{amssymb}
\usepackage{graphicx}
\usepackage{algorithm}
\usepackage{algorithmic}
\usepackage{makecell}  % 导言区加这个
\usepackage{multirow}
\usepackage{float}
\usepackage{tabularx}
\usepackage{array}
\usepackage{booktabs}
\usepackage{graphicx}
\usepackage{adjustbox}
\usepackage{tabularx}      % 用于创建可自动换行和调整列宽的表格
\definecolor{darkred}{rgb}{0.8,0,0}

\usepackage{listings}      % 用于优雅地展示代码块
\usepackage{todonotes}
\usepackage{colortbl} % 用于表格行底色 (\rowcolor)
\usepackage{graphicx} % 用于插入 logo
\usepackage{booktabs} % 用于三线表
\usepackage{pifont}
\usepackage{xcolor}

\newcommand{\cmark}{\textcolor{green!50!black}{\small\ding{51}}}
\newcommand{\xmark}{\textcolor{red!70!black}{\ding{56}}}

\usepackage[utf8]{inputenc} % allow utf-8 input
\usepackage[T1]{fontenc}    % use 8-bit T1 fonts
\usepackage{hyperref}       % hyperlinks
\usepackage{url}            % simple URL typesetting
\usepackage{booktabs}       % professional-quality tables
\usepackage{amsfonts}       % blackboard math symbols
\usepackage{nicefrac}       % compact symbols for 1/2, etc.
\usepackage{microtype}      % microtypography
\usepackage{xcolor}         % colors

\usepackage{graphicx}
\usepackage{subcaption}
\usepackage{amsmath}
\usepackage{amssymb}
\usepackage{mathtools}
\usepackage{amsthm}
\usepackage{tikz}
\usetikzlibrary{arrows.meta, positioning, fit, calc}

\usepackage[capitalize,noabbrev]{cleveref}

\usepackage{xcolor}
\definecolor{ForestGreen}{RGB}{34,139,34}

\theoremstyle{plain}

\theoremstyle{definition}

\theoremstyle{remark}

\usepackage{tabularx}
\usepackage{array}
\usepackage{graphicx}
\usepackage{booktabs}
\definecolor{deeppurple}{RGB}{109, 58, 159}
\hypersetup{
    colorlinks=true,
    citecolor=deeppurple,
    linkcolor=deeppurple,
    urlcolor=deeppurple
}
\newcolumntype{M}[1]{>{\centering\arraybackslash}m{#1}}
\newcolumntype{C}{>{\centering\arraybackslash}X}
\title{SpatialWorld: Benchmarking Interactive Spatial Reasoning of Multimodal Agents in Real-World Tasks}
\vspace{-0.2cm}
\renewcommand\footnotemark{}
\author{
  Hongcheng Gao$^{*\dagger1}$, \   Hailong Qu$^{*2}$, \   Jingyi Tang$^{3}$, \   Jiahao Wang$^{4}$, \   Zihao Huang$^{5}$ \\[0.1cm]
  \textbf{Hengkang Qiao}$^{2}$, \   \textbf{Shihong Huang}$^{3}$, \   \textbf{Junming Yang}$^{6}$, \   \textbf{Yi Li}$^{1}$, \   \textbf{Hongyixuan Yuan}$^{2}$ \\[0.1cm]
  \textbf{Wenjie Li}$^{7}$, \   \textbf{Bohan Zeng}$^{3}$, \   \textbf{Wenbo Li}$^{8}$, \   \textbf{Bo Wang}$^{5}$, \   \textbf{Jianhui Liu}$^{9}$, \   \textbf{Olive Huang}$^{3}$ \\[0.1cm]
  \textbf{Haoyang Huang}$^{8}$, \   \textbf{Wentao Zhang}$^{3}$, \   \textbf{Guoqing Huang}$^{2}$, \   \textbf{Nan Duan}$^{8}$, \   \textbf{Yinpeng Dong}$^{\dagger1}$
  \thanks{\!\!$^*$Equal contribution. $^{\dagger}$Corresponding author.} \\
  \\
  \small $^{1}$Tsinghua University \quad \small $^{2}$Chongqing University \quad \small $^{3}$Peking University \quad \\[-0.05cm]
  \small $^{4}$Xi'an Jiaotong University \quad \small $^{5}$Beijing Institute of Technology \quad \small $^{6}$Southeast University \\[-0.05cm]
  \small $^{7}$Shanghai Jiao Tong University \quad \small $^{8}$Joy Future Academy \quad \small $^{9}$The University of Hong Kong
}
\begin{document}
\vspace{-0.6cm}

\maketitle
\vspace{-0.85cm}
\begin{center} \textcolor{deeppurple}{Project Page: \href{https://spatial-world.github.io}{https://spatial-world.github.io}}
\end{center}
\vspace{0.1cm}
\begin{abstract}

Spatial reasoning is a foundational capability for multimodal large language models (MLLMs) to perceive and operate within the physical world. However, existing benchmarks predominantly rely on passive evaluation (e.g., static VQA) or simulator-specific pipelines, failing to assess general interactive spatial understanding. We introduce \textsc{SpatialWorld}, a unified benchmark designed specifically for evaluating the interactive spatial understanding of multimodal agents in complex real-world tasks. Integrating eight heterogeneous simulation backends under a shared, simulator-agnostic protocol, \textsc{SpatialWorld} features 760 human-annotated tasks across diverse domains (e.g., household routines, travel, social collaboration). Agents must solve tasks under vision-only partial observability, actively gathering egocentric visual evidence and expressing decisions via a unified, text-based action interface native to MLLMs. For reliable evaluation, each task includes a human-validated initial state, a reference trajectory, and a terminal-state verifier. Evaluating 15 advanced agents reveals that robust spatial task solving remains challenging: the strongest model, GPT-5, achieves an average task success rate (TSR) of only 17.4\%, while the leading open-source model, Qwen-3.5, reaches 14.1\%. Further analysis exposes a clear mismatch between task success and execution efficiency, alongside substantial domain-specific performance variations. These bottlenecks in active exploration and long-horizon planning position \textsc{SpatialWorld} as a rigorous testbed for future spatial agents.

% Extensive evaluation further reveals that current multimodal agents remain far from robust 3D task solving.
\end{abstract}

\vspace{-0.2cm}\section{Introduction}
\vspace{-0.2cm}

Spatial reasoning is a foundational capability for multimodal large language models (MLLMs) to perceive, understand, and operate within the physical world~\citep{gpt5_2,claude_45_opus,gemini_3_pro}. Existing benchmarks for spatial reasoning predominantly adopt a passive evaluation paradigm, such as static Visual Question Answering (VQA)~\citep{johnson2017clevr,hudson2019gqa,du2024embspatial,wu2025spatialscore,ma2022sqa3d} or the understanding of pre-recorded videos~\citep{yang2025thinking,lin2025ost,wang2025site}. Although these tasks assess the basic understanding of models regarding spatial relations, object layouts, and scene structures, they struggle to capture the interactive and dynamic nature of spatial understanding in real-world environments. 
% In physical spaces, an agent typically cannot acquire complete information in a single instance; instead, it must navigate actively and observe continuously to gather visual evidence progressively, update spatial beliefs under partial observability~\citep{zhu2024llava,cai2025scaling,chen2024spatialvlm}, and plan subsequent actions accordingly. Therefore, evaluating the spatial reasoning capability of MLLMs should not focus solely on their recognition and description of static scenes; it must also examine their ability to acquire information actively, update their understanding of the environment, and accomplish tasks during interactions.
Since physical spaces are partially observable, agents cannot acquire complete information from a single view. Instead, they must actively navigate to gather progressive visual evidence, update spatial beliefs~\citep{zhu2024llava,cai2025scaling,chen2024spatialvlm}, and plan subsequent actions. Therefore, evaluating MLLM spatial reasoning must move beyond static scene recognition to assess their capacity for dynamic exploration and interactive task completion.

Existing embodied benchmarks~\citep{shridhar2020alfred,yang2025embodiedbench,cheng2025embodiedeval,li2024eai} provide important interactive testbeds for navigation, manipulation, and task execution, yet many of them are designed around simulator-specific embodiments~\citep{shridhar2020alfred,zhang2024vlabench,yang2025embodiedbench}, sensor assumptions~\citep{savva2019habitat,li2021igibson,li2024eai}, action interfaces~\citep{zhang2024vlabench,cheng2025embodiedeval,li2024eai,liu2025spatial}, or execution pipelines~\citep{shridhar2020alfred,zhang2024vlabench,yang2025embodiedbench}.
% Existing embodied benchmarks~\citep{shridhar2020alfred,zhang2024vlabench,yang2025embodiedbench,cheng2025embodiedeval,li2024eai} extend spatial evaluation from static scene understanding to interactive settings through navigation, manipulation, and long-horizon task execution. However, many of these benchmarks remain tied to simulator-specific embodiments~\citep{shridhar2020alfred,zhang2024vlabench,yang2025embodiedbench}, explicit sensor assumptions~\citep{savva2019habitat,li2021igibson,li2024eai}, environment-coupled action interfaces~\citep{zhang2024vlabench,cheng2025embodiedeval,li2024eai,liu2025spatial}, or bespoke execution pipelines~\citep{shridhar2020alfred,zhang2024vlabench,yang2025embodiedbench}.
This makes it difficult to determine whether task success reflects general interactive spatial reasoning or adaptation to a particular simulator or action space~\citep{sohn2025embodied4cmeasuringmattersembodied}. The rapid progress of general MLLMs therefore raises a different evaluation question: can an off-the-shelf multimodal model, without being trained for a specific simulator, solve spatial tasks through egocentric visual observation, language-grounded high-level decisions, and closed-loop interaction across heterogeneous 3D environments? Answering this question requires an evaluation regime with three key properties. First, agents should operate under vision-only partial observability, without relying on additional sensor inputs or privileged state information~\citep{savva2019habitat,li2021igibson}. Second, the action interface should be native to MLLMs: expressing high-level navigation, viewpoint, interaction, and task-control decisions through a text-based action space enables the model to decompose and solve complex real-world tasks via explicit chain-of-thought reasoning~\citep{liu2025spatialcot}. Third, the protocol should be simulator-agnostic, using a unified interaction interface across environments rather than action designs deeply coupled with a single backend~\citep{liu2025spatial,li2023m3dbench,gholami2025spatial}. Such a highly decoupled paradigm eliminates the interference of low-level simulator characteristics, thereby providing a rigorous and faithful evaluation of the model's capacity for active exploration and decision-making based solely on visual observations and instructions.

To realize this objective, we introduce \textsc{SpatialWorld}, a unified benchmark designed specifically for evaluating the interactive spatial understanding of multimodal agents in 3D environments. As summarized in Table~\ref{tab:stats}, \textsc{SpatialWorld} contains 760 human-annotated tasks that span household routines, work and study, entertainment, travel, social collaboration, and digital spatial games. These tasks are instantiated across eight simulation backends, including AI2-THOR \citep{kolve2017ai2}, ProcTHOR \citep{DBLP:conf/nips/DeitkeVHWESHKKM22}, VirtualHome \citep{DBLP:conf/cvpr/PuigRBLWF018}, CARLA \citep{dosovitskiy2017carla}, EmbodiedCity~\citep{gao2024embodiedcity}, their multi-agent variants, and lightweight environments for 3D games. \textsc{SpatialWorld} wraps these heterogeneous platforms into a unified end-to-end evaluation framework with shared interfaces for observation, action, and verification. This design allows \textsc{SpatialWorld} to diagnose failures across complementary forms of 3D reasoning, rather than reducing the performance of agents to a single score specific to a simulator. By abstracting away the underlying complexities of disparate simulators, this unified architecture enables us to rigorously assess spatial reasoning under constraints that closely mimic real-world interactions. In contrast to prior benchmarks \citep{xie2024osworld,cao2024spider2v,yang2025embodiedbench}, our framework ensures that the performance of agents is evaluated reliably based on perceptual constraints that match tasks in the real world. To guarantee this reliability, each task is paired with an initial-state configuration validated by human annotators, a reference trajectory, and a task-specific terminal-state verifier. These artifacts allow us to measure not only whether an agent reaches the goal state, but also whether this achievement is realized through an efficient and interpretable sequence of actions.

We conduct extensive experiments on \textsc{SpatialWorld} with fifteen advanced multimodal agents from both open-source and proprietary model families. Our systematic evaluation reveals three major findings. First, current agents remain far from reliable in solving 3D tasks: across the full benchmark, the strongest model, GPT-5, achieves an average TSR of only 17.4\%, while the best open-source model, Qwen-3.5-397B-A17B, reaches 14.1\%. Second, there is a clear mismatch between task success and execution efficiency: models with a higher TSR do not necessarily achieve higher efficiency, which suggests that success is often accompanied by redundant exploration or shortcuts dependent on the task. Third, the rankings of models vary substantially across domains: 
% Qwen-3.5-397B-A17B leads in work and physical entertainment tasks, GPT-5 performs best in daily household and social collaboration tasks, and Gemini-3.1-Pro achieves the highest scores in travel and digital games.
GPT-5 leads daily household, travel, and social collaboration tasks, Qwen-3.5-397B-A17B ties GPT-5 in Work \& Study and leads physical entertainment, and Gemini-3.1-Pro achieves the highest scores in digital games.
These findings demonstrate that \textsc{SpatialWorld} exposes multiple, separable bottlenecks in spatial reasoning, long-horizon planning, and action execution, rather than reducing the evaluation of 3D tasks to a single score on a leaderboard.

\vspace{-0.2cm}\section{\textsc{SpatialWorld} Benchmark}\vspace{-0.3cm}

\begin{figure}
\vspace{-0.4cm}
    \centering
    \includegraphics[width=0.95\linewidth]{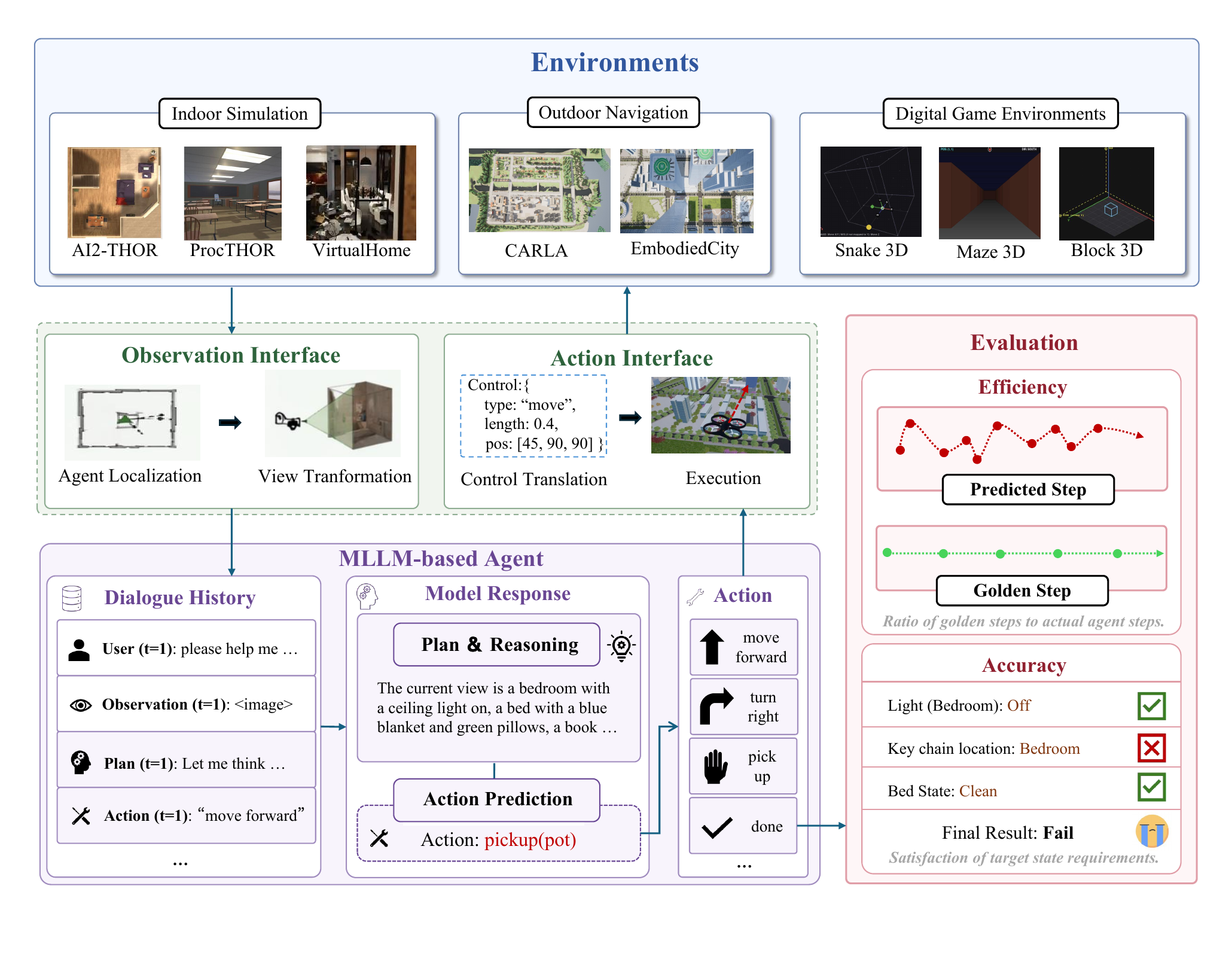}
    \vspace{-0.6cm}
    % \caption{SpatialWorld is a scalable, general-purpose evaluation framework for multimodal agents, supporting end-to-end task solving, structured plan generation. The framework unifies diverse 3D backends under a standardized observation and action interface. By facilitating complex, multi-goal missions, it enables rigorous assessment of an agent's spatial reasoning and long-horizon planning capabilities through reproducible benchmarks and automated efficiency metrics.}
    % \label{fig:placeholder}
    \caption{SpatialWorld is a scalable, general-purpose evaluation framework for multimodal agents, supporting end-to-end task solving and structured plan generation. It unifies diverse 3D backends under a standardized observation-action interface, enabling rigorous assessment of interactive spatial reasoning via reproducible benchmarks and automated efficiency metrics.}
    \label{fig:placeholder}
    \vspace{-0.2cm}
\end{figure}
 We formalize the task as a vision-only POMDP (\S\ref{sec:task_def}) and present \textsc{SpatialWorld}'s distinguishing design principles from existing benchmarks (\S\ref{sec:protocol}). We then describe the system architecture---a unified observation-action interface across heterogeneous simulators and execution-based evaluation (\S\ref{sec:obs_act})---followed by the construction pipeline covering task taxonomy and data annotation (\S\ref{sec:taxonomy}).

% \vspace{-0.2cm}\subsection{Task Definition}
% \label{sec:task_def}

% Each task in \textsc{SpatialWorld} is formalized as a partially observable Markov decision process (POMDP) $(S, O, A, T, R)$, where $S$ is the hidden environment state (object positions, physical properties, agent pose), $O$ is the egocentric visual observation space, $A$ is the high-level action space, $T$ is the simulator's transition function, and $R$ is the task-specific reward.
% At every step $t$, the agent receives a raw egocentric RGB screenshot $o_t \in O$ together with a natural-language task description $g$; no depth, semantic label, sensor reading, or global map is provided.
% The agent predicts an action $a_t \in A$, the simulator executes it and transitions to $s_{t+1}$, and a new observation $o_{t+1}$ is returned.
% A task terminates when the agent issues the \texttt{DONE} or \texttt{FAIL} signal, or when the step budget (default 40) is exhausted.
% This \emph{vision-only, multi-turn} formulation is the defining constraint of \textsc{SpatialWorld}: unlike VLA-style benchmarks that inject privileged state signals alongside visual input, agents here must reason purely from egocentric images---the same perceptual conditions a human operator faces.

\vspace{-0.2cm}\subsection{Task Formulation}\vspace{-0.2cm}
\label{sec:task_def}

Each task in \textsc{SpatialWorld} is formulated as a partially observable Markov decision process (POMDP) defined by the tuple $\langle \mathcal{S}, \mathcal{O}, \mathcal{A}, \mathcal{T}, \Omega, \mathcal{R} \rangle$. At step $t$, the agent receives a natural-language goal $g$ and a raw egocentric RGB observation $o_t \in \mathcal{O}$, strictly without privileged state signals (e.g., depth or global maps). Given the trajectory history $\mathcal{H}_t = (o_1, a_1, \dots, o_t)$, the MLLM-based policy $\pi_\theta$ predicts the next high-level action $a_t \sim \pi_\theta(a_t \mid \mathcal{H}_t, g)\in \mathcal{A}$. The simulator then executes $a_t$, transitions the hidden environment state $s_{t+1}$, and renders the next visual observation $o_{t+1}$:
% \vspace{-0.1cm}
\begin{equation}
    s_{t+1} \sim \mathcal{T}(s_{t+1} \mid s_t, a_t), \quad o_{t+1} \sim \Omega(o_{t+1} \mid s_{t+1}).
\end{equation}
% \vspace{-0.1cm}
This dynamic interaction continues until the agent executes \texttt{EndTask} or exhausts the step budget. By enforcing this strict \emph{vision-only, multi-turn} formulation, \textsc{SpatialWorld} ensures agents must reason under the exact same perceptual conditions as a human operator, distinguishing it from benchmarks that rely on privileged state inputs.

\begin{figure}
\vspace{-0.4cm}
    \centering
    \includegraphics[width=0.92\linewidth, height=9.8cm, keepaspectratio=false]{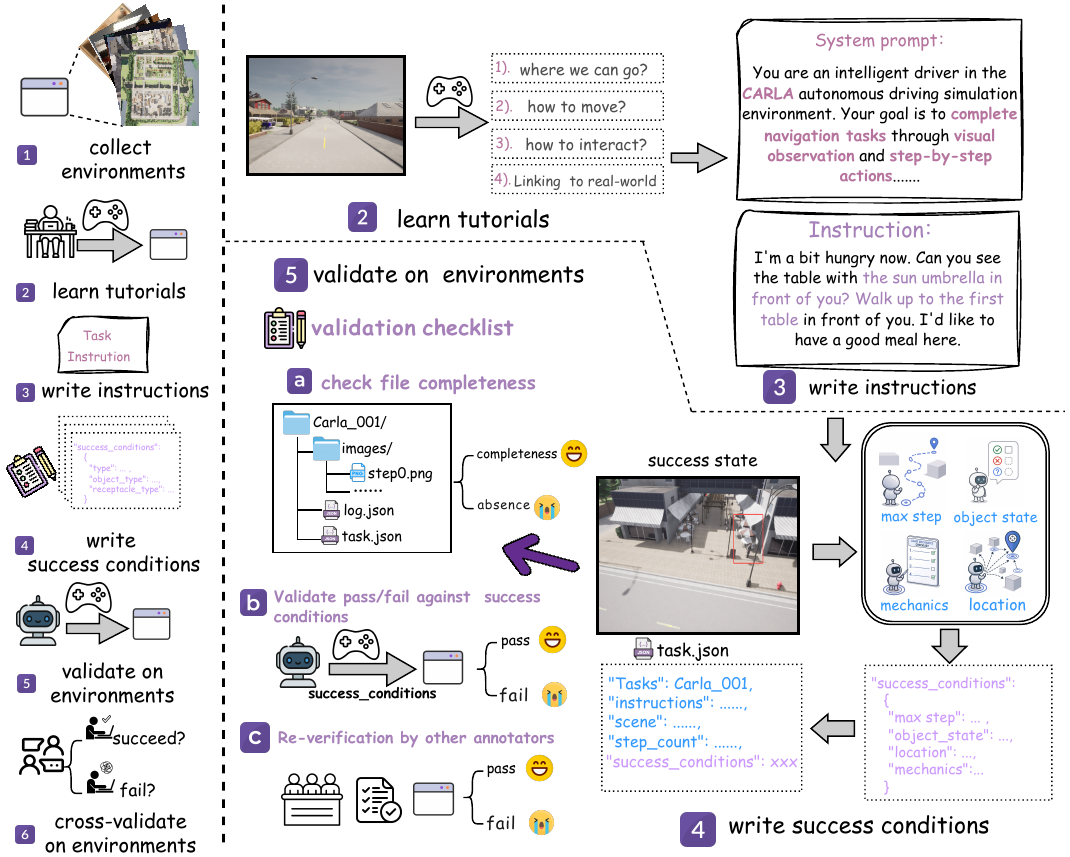}
    % \caption{\textbf{Data construction pipeline of SpatialWorld.} We adopt this pipeline across all environments, and have specifically trained annotators and hired expert teams to assist with annotation and verification. }
    \caption{\textbf{Data construction pipeline of SpatialWorld.} We first collect a series of environments, have annotators learn tutorials and write instructions, define success conditions, and then calibrate the data through automated execution validation in virtual environments and human cross-validation.}
    % We provide example figures for all environments in Fig.~\ref{fig:placeholder} and include detailed environment introduction documents in the open-source code.
    \label{fig:pipeline}
    \vspace{-0.25cm}
\end{figure}

\vspace{-0.1cm}\subsection{Benchmark Protocol}
\label{sec:protocol}
\begin{table*}[t]
\centering
\small
% \caption{\textbf{Spatial benchmark comparison.} Representative spatial ImageQA, VideoQA, and embodied-agent benchmarks differ in the properties that motivate \textsc{SpatialWorld}'s construction: a unified cross-platform interface, interactivity, first-person observations, vision-only inputs, and language-form outputs. Instances reports the original number of evaluation instances stated by each benchmark; \cmark{} and \xmark{} indicate presence and absence, respectively. More comparison is provided in Appendix~\ref{sec:app_benchmark_compare}.}
\caption{\textbf{Spatial benchmark comparison.} Representative spatial ImageQA, VideoQA, and embodied-agent benchmarks differ in key properties motivating \textsc{SpatialWorld}: unified cross-platform interface, interactivity, first-person observations, vision-only inputs, and language-form outputs. \cmark{} and \xmark{} indicate presence and absence. Further comparison in Appendix~\ref{sec:app_benchmark_compare}.}
\vspace{-0.1cm}
\label{tab:benchmark_compare}
\resizebox{0.98\textwidth}{!}{%
\begin{tabular}{llcccccc}
\toprule
\textbf{Type} & \textbf{Benchmark} & \textbf{Instances} & \makecell[c]{\textbf{Unified cross-}\\\textbf{platform interface}} & \makecell[c]{\textbf{Interactive}\\\textbf{env.}} & \makecell[c]{\textbf{First-person}\\\textbf{observation}} & \makecell[c]{\textbf{Vision-only}\\\textbf{input}} & \makecell[c]{\textbf{Language-form}\\\textbf{output}} \\
\midrule

% \toprule
% \shortstack{\textbf{Task}\\\textbf{Type}} & \shortstack{\textbf{Benchmark}\\\textbf{Name}} & \shortstack{\textbf{Eval.}\\\textbf{Instances}} & \shortstack{\textbf{Unified cross-}\\\textbf{platform interface}} & \shortstack{\textbf{Interactive}\\\textbf{env.}} & \shortstack{\textbf{First-person}\\\textbf{observation}} & \shortstack{\textbf{Vision-only}\\\textbf{input}} & \shortstack{\textbf{Language-form}\\\textbf{output}} \\
% \midrule
\multirow{4}{*}{ImageQA} 
& SpatialEval~\citep{wang2024spatialeval} & 4635 & \xmark & \xmark & \xmark & \cmark & \cmark \\
& 3DSRBench~\citep{ma20253dsrbench} & 2772 & \xmark & \xmark & \xmark & \cmark & \cmark \\
& EmbSpatial-Bench~\citep{du2024embspatial} & 3640 & \xmark & \xmark & \cmark & \cmark & \cmark \\
& SpatialScore~\citep{wu2025spatialscore} & 5025 & \xmark & \xmark & \xmark & \cmark & \cmark \\
\midrule
\multirow{3}{*}{VideoQA}
& SpatialBench~\citep{xu2025spatialbench} & 3193 & \xmark & \xmark & \cmark & \cmark & \cmark \\
& SITE~\citep{wang2025site} & 8068 & \xmark & \xmark & \xmark & \cmark & \cmark \\
& VSI-Bench~\citep{yang2025thinking} & 5130 & \xmark & \xmark & \cmark & \cmark & \cmark \\
% & OST-Bench~\citep{lin2025ost} & 10k & \xmark & \xmark & \cmark & \cmark & \cmark \\
\midrule
\multirow{2}{*}{Embodied Bench}
& ALFRED~\citep{shridhar2020alfred} & 25.7k & \xmark & \cmark & \cmark & \cmark & \xmark \\
& VLABench~\citep{zhang2024vlabench} & 100 & \xmark & \cmark & \cmark & \xmark & \xmark \\
% & EmbodiedBench~\citep{yang2025embodiedbench} & 1128 & \cmark & \cmark & \cmark & \xmark & \cmark \\
% & EmbodiedCity~\citep{gao2024embodiedcity} & 87.1k & \xmark & \cmark & \cmark & \cmark & \cmark \\
\midrule
\textbf{Ours}
& \textbf{\textsc{SpatialWorld}} & \textbf{760} & \cmark & \cmark & \cmark & \cmark & \cmark \\
\bottomrule
\end{tabular}%
}
\vspace{-0.5cm}
\end{table*}

To rigorously evaluate the active spatial reasoning capabilities of general MLLMs, we introduce \textsc{SpatialWorld}. As shown in Table~\ref{tab:benchmark_compare}, existing spatial benchmarks typically fall into two categories: static ImageQA/VideoQA datasets that fail to capture dynamic environmental interactions, and simulator-coupled embodied frameworks that rely heavily on non-visual metadata and low-level action parameters. Addressing these limitations, \textsc{SpatialWorld} introduces a unified, interactive evaluation paradigm guided by four core design principles:
\textbf{(1) Pure egocentric vision:} As highlighted in Table~\ref{tab:benchmark_compare}, agents receive strictly first-person, vision-only observations without any privileged state information (e.g., ground-truth object coordinates or semantic metadata). This guarantees a genuine evaluation of the coupling between visual perception and spatial reasoning. \textbf{(2) Cross-platform unification:} We abstract simulator-specific complexities into a unified interface driven by standardized language-form outputs. This enables the direct evaluation of general MLLMs across distinct domains while fully preserving the native physical challenges of each underlying environment. \textbf{(3) Factored complexity:} We systematically decouple photorealistic visual semantics from pure geometric reasoning by incorporating abstract 3D games alongside daily embodied setups. This factorization allows us to pinpoint specific bottlenecks in a model's spatial cognition without confounding variables. \textbf{(4) Execution-based verification:} Success is objectively verified via terminal environment states rather than strict adherence to predefined action trajectories. This approach accommodates the open-ended exploration and diverse reasoning paths characteristic of autonomous MLLM agents.

\vspace{-0.2cm}\subsection{SpatialWorld Architecture}
\vspace{-0.2cm}
\label{sec:obs_act}

% As illustrated in Fig.~\ref{fig:placeholder}, the architecture of Spatial World consists of five primary components from a macroscopic perspective:
% (i) The \textbf{Environment Interface} is responsible for constructing the specific environments corresponding to different tasks.
% (ii) The \textbf{Observation Interface} extracts visual observations from the instantiated environment.
% (iii) The \textbf{MLLM-based Agent Module} receives these environmental observations, performs reasoning and planning, and outputs corresponding decision-making actions.
% (iv) The \textbf{Action Interface} translates the model's unified output into environment-specific action commands.
% (v) The \textbf{Verification Interface} evaluates whether the agent has successfully executed the task upon termination of the interaction or when the maximum number of interaction steps is reached.

\begin{figure}
     \vspace{-.6cm}
    \centering
    \includegraphics[width=0.99\linewidth]{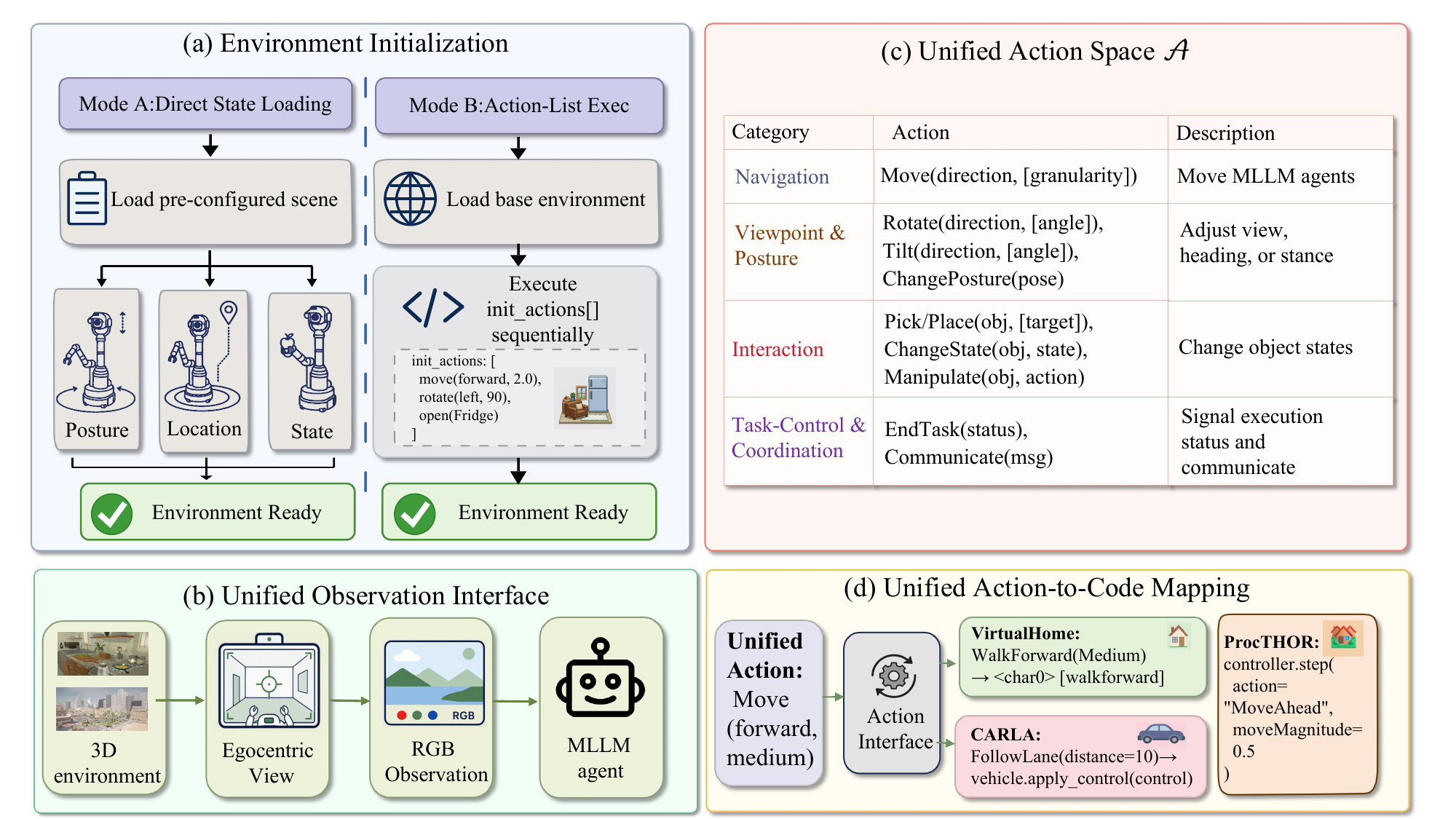}
    % \caption{\textbf{The Observation and Action Interfaces of \textsc{SpatialWorld}}. (a) SpatialWorld supports two initialization modes: Direct State Loading instantiates a fully specified scene configuration, while Action-List Execution starts from a base environment and replays setup actions to reach the desired initial state.  (b) The observation interface normalizes heterogeneous 3D environments into egocentric RGB observations for the MLLM agent.  (c) The unified action space $\mathcal{A}$ organizes embodied actions into Navigation, Viewpoint \& Posture, Interaction, and Task-Control \& Coordination categories.  (d) The Action Interface translates each unified action into environment-specific execution code, enabling cross-simulator deployment across VirtualHome, CARLA, and ProcTHOR.}
        % \vspace{-.1cm}
    \caption{\textbf{The Observation and Action Interfaces.} (a) Flexible environment initialization via direct state loading or action-list execution. (b) A unified interface providing standardized egocentric RGB observations. (c) A structured, unified action space $\mathcal{A}$. (d) Action-to-code mapping that translates unified actions into environment-specific commands, enabling cross-simulator deployment.}
    \label{fig:action_architecture}
    \vspace{-.6cm}
\end{figure}

To systematically decouple environment execution from agent decision-making, \textsc{SpatialWorld} introduces a modular architecture comprising five standardized components, as illustrated in Fig.~\ref{fig:placeholder}. 
In this closed-loop pipeline, the \textbf{Environment} and \textbf{Verification} interfaces rigorously manage task initialization and deterministic success checking, while the \textbf{Agent Module} acts as the central MLLM-based reasoning engine. 
Crucially, to bridge the gap between diverse simulator backends and a unified agent policy, the \textbf{Observation} and \textbf{Action} interfaces are designed as a strictly defined I/O bottleneck: they encapsulate all heterogeneous sensory rendering and physics-engine executions into a standardized interaction protocol. 
As illustrated in Fig.~\ref{fig:action_architecture}, these interfaces serve as a crucial bridge by normalizing complex 3D environments into egocentric visual observations and translating high-level decisions into simulator-specific execution codes. Guided by this architectural abstraction, we next formally detail the unified observation and action spaces exposed to the agent.

%这里不起章节，另起一句话，然后引出observation and action space，然后注意提一下接口，引出space，因一下 Figure4，如什么图4XXX所言。

%jingyi:这个地方需要总结一下interface，然后引出下面的space，然后印一下图4
\vspace{-0.1cm}
\textbf{Observation Space.}
At each step, the agent receives a single egocentric RGB screenshot at the simulator's native resolution.
No auxiliary modality (e.g., depth map, optical flow, semantic segmentation, or global occupancy map) is available.
This vision-only constraint is the primary departure from VLA-style benchmarks, which typically inject privileged sensor states (joint angles, object lists, navigation graphs) alongside visual input, and from offline 3D benchmarks, which supply pre-captured multi-view scans or videos rather than requiring active information gathering.

\vspace{-0.1cm}
\textbf{Action Space.}
Rather than requiring low-level continuous motor commands (e.g., joint torques or velocity vectors), we expose a unified high-level action space $\mathcal{A}$ that abstracts heterogeneous simulator backends behind a common symbolic interface.
This design choice serves two purposes: (i) it enables direct evaluation of off-the-shelf MLLMs without task-specific fine-tuning, and (ii) it produces interpretable, language-grounded reasoning traces as a natural byproduct of decision-making.
% Concretely, $\mathcal{A}$ is organized into four functional categories: (i)~\emph{locomotion} primitives (\texttt{move}) for spatial displacement, (ii)~\emph{viewpoint} primitives (\texttt{rotate}) for reorienting the agent's egocentric perspective, (iii)~\emph{manipulation} primitives (\texttt{pick}, \texttt{place}, \texttt{open}, \texttt{close}, \texttt{toggle}, \texttt{break}, \texttt{fill}) for object-state interaction, and (iv)~\emph{meta-control} signals (\texttt{WAIT}, \texttt{DONE}, \texttt{FAIL}) for execution pacing and termination.

% Concretely, $\mathcal{A}$ follows the four functional categories shown in Fig.~\ref{fig:action_architecture}: (i)~\emph{Navigation} actions such as \texttt{Move(direction, [granularity])} for spatial displacement, (ii)~\emph{Viewpoint \& Posture} actions such as \texttt{Rotate(direction, [angle])}, \texttt{Tilt(direction, [angle])}, and \texttt{ChangePosture(pose)} for adjusting heading, camera angle, or stance, (iii)~\emph{Interaction} actions such as \texttt{Pick/Place(obj, [target])}, \texttt{ChangeState(obj, state)}, and \texttt{Manipulate(obj, action)} for object-level changes, and (iv)~\emph{Task-Control \& Coordination} actions such as \texttt{EndTask(status)} and \texttt{Communicate(msg)} for termination, pacing, and multi-agent communication.
Concretely, as shown in Fig.~\ref{fig:action_architecture}, $\mathcal{A}$ encompasses four high-level functional categories: (i)~\emph{Navigation} (e.g., \texttt{Move}), (ii)~\emph{Viewpoint \& Posture} (e.g., \texttt{Rotate}), (iii)~\emph{Interaction} (e.g., \texttt{Pick/Place}), and (iv)~\emph{Task-Control \& Coordination} (e.g., \texttt{EndTask}). 
The Action Interface translates these unified primitives into simulator-specific execution calls, ensuring that a single agent policy generalizes across diverse environments without modification. Detailed definitions of all available actions and parameters are deferred to Appendix~\ref{sec:appendix_action}.
The Action Interface (component iv) maps these unified primitives to simulator-specific execution calls, ensuring that a single agent policy generalizes across environments without modification (see Fig.~\ref{fig:action_architecture} for a complete overview).

%jingyi:这个部分把backup的内容整理过来缩写，保持段落长度
% \textbf{Evaluation Metrics.}
% The Verification Interface (component v) assesses task completion through two complementary metrics.
% The primary metric is the \textbf{TSR}, which measures the fraction of tasks where the agent fully satisfies the terminal goal condition:
% \begin{equation}
% \text{TSR} = \frac{1}{N}\sum_{i=1}^{N} \mathbf{1}\!\left[\mathcal{V}_i(s_T^{(i)}) = 1\right],
% \end{equation}
% where $s_T^{(i)}$ is the terminal environment state for task $i$ and $\mathcal{V}_i$ is the task-specific verifier. For Snake3D, exact completion is too sparse to separate weak partial progress from complete failure, so we report a scale-normalized discrete score by dividing the achieved snake score by the spatial edge length of the game environment.
% Beyond binary success, we also measure \textbf{Step Efficiency (SE)} to capture how efficiently an agent completes a task.
% For each successfully completed task $i$, let $n_i$ denote the number of steps taken and $n_i^*$ the number of steps in the human-annotated reference solution:
% \begin{equation}
% \text{SE} = \frac{1}{|\mathcal{S}|}\sum_{i \in \mathcal{S}} \frac{n_i^*}{n_i},
% \end{equation}
% where $\mathcal{S}$ is the set of successfully completed tasks.
% An SE of 1.0 indicates the agent matches the reference solution length exactly; lower values indicate unnecessary steps.
% Reporting both TSR and SE jointly distinguishes agents that succeed efficiently from those that succeed by exhaustive trial-and-error within the step budget.
\textbf{Environment Suite.} We integrate eight backends under a unified agent-side abstraction to ensure cross-environment comparisons evaluate genuine \emph{real-world interactive spatial understanding} rather than interface bias. The suite is organized into three families: \emph{Indoor Simulation} (AI2-THOR~\cite{kolve2017ai2}, ProcTHOR~\cite{DBLP:conf/nips/DeitkeVHWESHKKM22}, VirtualHome~\cite{DBLP:conf/cvpr/PuigRBLWF018}) provides explicit physical affordances to test fine-grained object grounding, temporally ordered routines, and multi-agent coordination. \emph{Outdoor Navigation} (CARLA~\cite{dosovitskiy2017carla}, EmbodiedCity) extends to macroscopic scales, evaluating long-range route planning and progress estimation across dynamic urban and aerial topologies. Finally, while realistic simulators are indispensable, their difficulty is often entangled with photorealistic semantics and natural scene priors. To address this, we specifically implemented \emph{Custom Digital Games} (e.g., Block3D, Snake3D, Rubik’s Cube) as controlled closed-loop probes. By stripping away visual shortcuts, these lightweight environments isolate the abstract spatial logic and topological reasoning that fundamentally underpin real-world interactive spatial understanding. Detailed descriptions are provided in Appendix~\ref{sec:app_env_suite}.

\textbf{Execution-Based Evaluation.}
Following OSWorld~\cite{xie2024osworld} and Spider2-V~\cite{cao2024spider2v}, \textsc{SpatialWorld} adopts terminal-state verification rather than static trajectory matching. A custom verifier $\mathcal{V}_i$ queries the final state to assess performance via two complementary metrics.
The primary metric is \textbf{TSR}, which measures the fraction of tasks where the terminal goal is fully satisfied: $\text{TSR} = \frac{1}{N}\sum_{i=1}^{N} \mathbf{1}[\mathcal{V}_i(s_T^{(i)}) = 1]$, where $s_T^{(i)}$ is the terminal state for task $i$ (task-specific adaptations, e.g., for Snake3D, are detailed in Appendix~\ref{sec:eval_details}).
To evaluate efficiency beyond success, we measure \textbf{Step Efficiency (SE)}: $\text{SE} = \frac{1}{|\mathcal{S}|}\sum_{i \in \mathcal{S}} \frac{n_i^*}{n_i}$, where $\mathcal{S}$ is the set of successful tasks, $n_i$ is the agent's step count, and $n_i^*$ is the human-annotated reference length. Jointly reporting TSR and SE distinguishes efficient agents from exhaustive trial-and-error.

\vspace{-0.2cm}\subsection{Benchmark Construction} \vspace{-0.2cm}
\label{sec:taxonomy}

%jingyi:这里需要修改下，这里的逻辑有问题，而且阐述不准确。
To systematically standardize the evaluation of spatial intelligence, \textsc{SpatialWorld} establishes a comprehensive benchmark construction protocol for its 760 tasks. Specifically, we first define a rigorous task taxonomy structured along two core dimensions: \textbf{Scenario Categories} and \textbf{Complexity Levels}, ensuring a diverse and hierarchical coverage of spatial capabilities. Guided by this taxonomy, we then employ a unified \textbf{Data Construction} pipeline to guarantee the high quality, consistency, and reproducibility of the entire dataset.

\textbf{Scenario Categories.}
\textsc{SpatialWorld} separates everyday embodied operation from abstract spatial reasoning while keeping both under the same closed-loop evaluation protocol.
The physical portion covers household routines, study and work activities, entertainment scenarios, travel-oriented navigation, and social collaboration, so that the benchmark spans object-centric manipulation, room-scale exploration, large-scale movement, and multi-agent coordination rather than a single simulator-specific skill.
The digital portion introduces 3D games as a complementary family: these tasks remove photorealistic semantics and instead emphasize geometric counting, maze planning, state tracking, and spatial transformation.
Table~\ref{tab:stats} summarizes how these scenario categories are distributed across the eight environments.

\begin{table*}[t]
\centering 

% ================= 左侧：表格区域 =================
\begin{minipage}[c]{0.58\textwidth}
\vspace{-0.4cm}
    \captionof{table}{\textbf{Scenario distribution.} Task distribution across environments and scenario categories. ``Social'' denotes Social Collaboration; ``Entertain.'' denotes Entertainment.}
    \vspace{-0.25cm}
    \label{tab:stats}
    \vspace{0.05in} % 表格和顶部标题的间距，可酌情调小
    \resizebox{0.97\linewidth}{!}{%
    \begin{tabular}{lcccccc}
\toprule
\textbf{Environment} & \textbf{Daily} & \textbf{Work} & \textbf{Entertain.} & \textbf{Travel} & \textbf{Social} & \textbf{Total} \\
\midrule
AI2-THOR        & 219 & 41 & 40 & 11 & 0 & 311 \\
ProcTHOR        & 92 & 10 & 23 & 2 & 0 & 127 \\
VirtualHome     & 27 & 8 & 3 & 0 & 0 & 38 \\
CARLA           & 0 & 0 & 0 & 80 & 0 & 80 \\
EmbodiedCity    & 12 & 0 & 2 & 39 & 0 & 53 \\
Multi-AI2THOR   & 0 & 0 & 0 & 0 & 29 & 29 \\
Multi-ProcTHOR  & 0 & 0 & 0 & 0 & 17 & 17 \\
3D Games        & 0 & 0 & 105 & 0 & 0 & 105 \\
\midrule
\textbf{Total} & \textbf{350} & \textbf{59} & \textbf{173} & \textbf{132} & \textbf{46} & \textbf{760} \\
\bottomrule
\end{tabular}
    }
\end{minipage}\hfill
% ================= 右侧：图片区域 =================
\begin{minipage}[c]{0.39\textwidth}
    \centering
    \vspace{-0.28cm}
    \includegraphics[width=0.97\linewidth]{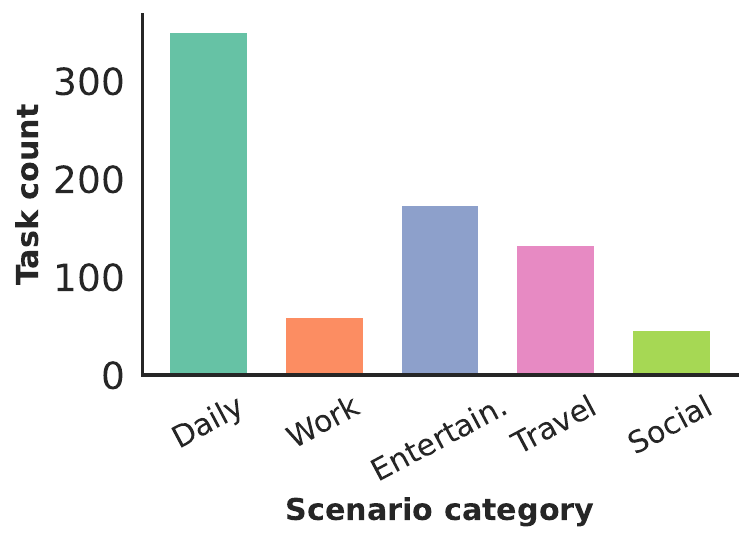}
    
    % 关键修改：使用负数间距将 caption 向上拉
    \vspace{-0.1in} % 如果还是觉得宽，可以改成 -0.2in；如果觉得太挤了，改成 -0.05in
    % \vspace{-0.1cm}
    \captionof{figure}{\textbf{Task-category counts.} Task distribution across different categories.}
    \label{fig:task_dist}
\end{minipage}
\vspace{-0.6cm}
\end{table*}

%jingyi:后面加分析，然后这里L1、L2、L3不是递进关系而是并列关系。
%根据这个label在后面加一个分析实验。
\textbf{Complexity Levels.}
Each task carries one of three complexity labels that reflect the cognitive demands placed on the agent.
\textbf{Navigation} tasks require the agent to explore the 3D environment and reach a target location or object, without manipulating environment state.
\textbf{Interaction} tasks require object-level state changes---picking, placing, opening, or toggling objects---but do not demand extensive spatial exploration.
\textbf{Hybrid} tasks combine long-horizon navigation with multi-step manipulation, demanding both spatial exploration and fine-grained physical interaction.

\textbf{Data Construction.}
As illustrated in Fig.~\ref{fig:pipeline}, we adopt a unified data construction pipeline across all environments in \textsc{SpatialWorld}.
For each task, the construction process consists of three stages: (i)~\emph{task design}, where annotators define the natural-language instruction and configure the initial environment state; (ii)~\emph{human execution}, where trained annotators independently solve each task in the simulator, recording the ground-truth terminal state and reference action sequence; and (iii)~\emph{verification}, where separate expert reviewers cross-check the task feasibility, instruction clarity, and evaluation script correctness.
All verifier logic and success conditions are validated through rigorous inter-annotator cross-checking, further ensuring the consistency, accuracy, reproducibility, and unambiguity of the evaluation signal.
Representative examples of the annotated evaluation scripts and human-validated cases are provided in Appendix~\ref{sec:app_human_annotation}.

\vspace{-0.2cm}\section{Experiment}
\vspace{-0.2cm}\subsection{Experimental Setup} \vspace{-0.2cm}
\label{sec:eval}

\textbf{Models and Tasks.}
We benchmark 15 state-of-the-art MLLMs spanning open-source and proprietary families:
\emph{Qwen series}~\citep{bai2025qwen25vltechnicalreport, yang2025qwen3technicalreport, qwen35blog};
\emph{GLM series}~\citep{hong2025glm, Glm4p6v};
\emph{Kimi series}~\citep{team2025kimi, team2026kimi};
\emph{Gemini series}~\citep{comanici2025gemini25pushingfrontier, gemini3, gemini3flash};
\emph{GPT series}~\citep{singh2025openai, gpt5p4};
\emph{Seed Series}~\citep{seed}.
% \textbf{Anthropic}---Claude-Sonnet-4.6.
All models are evaluated using their official APIs or open-weight checkpoints without task-specific fine-tuning.
Each model is prompted with the egocentric RGB screenshot and a natural-language task description at every step; no privileged state information is provided. All models are evaluated on the full \textsc{SpatialWorld} benchmark comprising 760 tasks across 8 simulation environments.

% the latest 50 turns of interaction
% \textbf{Evaluation Details.}
\textbf{Evaluation Details.}
We use temperature $\tau = 1.0$ and retain the latest $w = 30$ turns of interaction as context for all main experiments.
The step budget for each task is dynamically determined as $2g + 10$, where $g$ denotes the golden action count annotated by human annotators.
Unless otherwise specified, we report TSR and SE aggregated over evaluated trajectories.

\vspace{-0.2cm}\subsection{Main Results}

\begin{table*}[t]
\vspace{-0.2cm}
\centering
\small
% \caption{\textbf{Performance Evaluation.} Main-benchmark TSR (\%) across task categories for the 15 evaluated models. Bold and underlined entries denote the best and second-best values in each column, respectively. Physical categories are derived from the benchmark scenario taxonomy; the digital category corresponds to the 3D game suite. Physical Overall is the average of Daily, Work, Entertain., Travel, and Social categories.}
\caption{\textbf{Performance Evaluation.} Main-benchmark TSR (\%) across task categories for 15 evaluated models. Bold and underlined entries denote the best and second-best per column. Physical categories follow the benchmark scenario taxonomy; digital corresponds to the 3D game suite. Physical Overall is the weighted average of Daily, Work, Entertain., Travel, and Social categories.}
\label{tab:main}
\vspace{-0.1cm}
\renewcommand{\arraystretch}{0.9}
\resizebox{0.96\textwidth}{!}{%
\begin{tabular}{lccccccc}
\toprule
\multirow{2}{*}{\textbf{Model}} & \multicolumn{6}{c}{\textbf{Physical}} & \multicolumn{1}{c}{\textbf{Digital}} \\
\cmidrule(lr){2-7}\cmidrule(lr){8-8}
 & \textbf{Daily} & \textbf{Work} & \textbf{Entertain.} & \textbf{Travel} & \textbf{Social} & \textbf{Overall} & \textbf{Entertain.} \\
\midrule
\rowcolor{purple!15} \multicolumn{8}{c}{\textbf{\textit{(A) Open-Source Models}}} \\
\midrule
\raisebox{-0.2ex}{\includegraphics[height=1em]{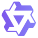}}~Qwen2.5-VL-72B~\citep{bai2025qwen25vltechnicalreport} & 3.7 & 8.5 & 2.9 & 0.8 & 2.2 & 3.4 & 7.6 \\
\raisebox{-0.2ex}{\includegraphics[height=1em]{logos/qwen-color.pdf}}~Qwen3-VL-30B-A3B~\citep{yang2025qwen3technicalreport} & 6.3 & 5.1 & 4.4 & 1.5 & 4.3 & 4.9 & 7.9 \\
\raisebox{-0.2ex}{\includegraphics[height=1em]{logos/qwen-color.pdf}}~Qwen3-VL-235B-Instruct~\citep{yang2025qwen3technicalreport} & 6.9 & 8.5 & 7.4 & 4.5 & 10.9 & 6.9 & 5.0 \\
\raisebox{-0.2ex}{\includegraphics[height=1em]{logos/qwen-color.pdf}}~Qwen3-VL-235B-Thinking~\citep{yang2025qwen3technicalreport} & 5.7 & 8.5 & 7.4 & 3.8 & 10.9 & 6.1 & 28.3 \\
\raisebox{-0.2ex}{\includegraphics[height=1em]{logos/qwen-color.pdf}}~Qwen-3.5-397B-A17B~\citep{qwen35blog} & \underline{13.1} & \textbf{16.9} & \textbf{13.2} & 4.5 & \underline{19.6} & \underline{12.2} & 26.0 \\
\raisebox{-0.2ex}{\includegraphics[height=1em]{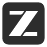}}~GLM-4.5V~\citep{hong2025glm} & 3.7 & 3.4 & 4.4 & 1.5 & 13.0 & 4.0 & 14.5 \\
\raisebox{-0.2ex}{\includegraphics[height=1em]{logos/glm-color.pdf}}~GLM-4.6V~\citep{Glm4p6v} & 2.9 & 5.1 & 4.4 & 1.5 & 0.0 & 2.7 & 8.1 \\
\raisebox{-0.2ex}{\includegraphics[height=1em]{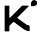}}~Kimi-VL-A3B~\citep{team2025kimi} & 1.1 & 3.4 & 0.0 & 0.0 & 0.0 & 0.9 & 3.3 \\
\raisebox{-0.2ex}{\includegraphics[height=1em]{logos/kimi.pdf}}~Kimi-K2.5~\citep{team2026kimi} & 11.1 & 8.5 & 4.4 & 3.8 & 17.4 & 9.2 & 31.0 \\

\midrule
\rowcolor{blue!15} \multicolumn{8}{c}{\textbf{\textit{(B) Closed-Source Models}}} \\
\midrule
\raisebox{-0.2ex}{\includegraphics[height=1em]{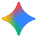}}~Gemini-2.5-Pro~\citep{comanici2025gemini25pushingfrontier} & 7.4 & \underline{11.9} & 1.5 & 3.8 & 10.9 & 6.7 & 32.6 \\
\raisebox{-0.2ex}{\includegraphics[height=1em]{logos/gemini-color.pdf}}~Gemini-3-Flash~\citep{gemini3flash} & 8.0 & 10.2 & 4.4 & \underline{6.1} & 4.3 & 7.2 & \underline{38.1} \\
\raisebox{-0.2ex}{\includegraphics[height=1em]{logos/gemini-color.pdf}}~Gemini-3.1-Pro~\citep{gemini3} & 11.4 & 10.2 & 5.9 & 4.5 & 8.7 & 9.2 & \textbf{39.0} \\
\raisebox{-0.2ex}{\includegraphics[height=1em]{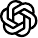}}~GPT-5~\citep{singh2025openai} & \textbf{14.9} & \textbf{16.9} & \underline{10.3} & \textbf{6.8} & \textbf{34.8} & \textbf{14.4} & 36.4 \\
\raisebox{-0.2ex}{\includegraphics[height=1em]{logos/openai.pdf}}~GPT-5.4~\citep{gpt5p4} & 8.0 & 5.1 & 5.9 & 3.8 & 6.5 & 6.6 & 11.9 \\
\raisebox{-0.2ex}{\includegraphics[height=1em]{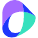}}~Doubao-2.0-Lite~\citep{seed} & 5.7 & 6.8 & 5.9 & 3.0 & 13.0 & 5.8 & 24.8 \\
\bottomrule
\end{tabular}%
}
\vspace{-0.2cm}
\end{table*}

Table~\ref{tab:main} and Table~\ref{tab:main_se} benchmark the TSR and SE performance of state-of-the-art MLLMs, revealing the following key insights:

\noindent\textbf{A significant gap remains between MLLM agents and real-world 3D environments.} Current models struggle significantly with physical tasks, where the best-performing 
% Qwen-3.5-397B-A17B and GPT-5 tie for the highest overall TSR at only 26.5\%.
GPT-5 reaches only 14.4\% Physical Overall TSR, followed by Qwen-3.5-397B-A17B at 12.2\%.
Furthermore, these successes are predominantly restricted to short-horizon, fundamental operations (e.g., turning on a device). Despite moderately better performance in digital domains, the universally low success rates underscore a persistent shortfall in human-level spatial intelligence.

\noindent\textbf{SE reveals reliance on trial-and-error among similarly capable models.} 
% SE is particularly informative for differentiating models with comparable Task Success Rates (TSR). 
% SE is particularly informative for differentiating models with comparable TSR; across large TSR gaps, task difficulty distributions may confound the comparison.
SE is informative for differentiating models with comparable TSR; large TSR gaps render it incomparable due to divergent completed task counts and difficulty distributions.
For instance, 
% Kimi-K2.5 and GPT-5.4 exhibit similar Physical Overall TSRs (19.4\% vs. 18.2\%), yet GPT-5.4 achieves a significantly higher SE (0.690 vs. 0.509)
Kimi-K2.5 and GPT-5.4 exhibit comparable Physical Overall TSRs (9.2\% vs. 6.6\%), yet GPT-5.4 achieves a higher SE (0.745 vs. 0.584). This contrast indicates that Kimi-K2.5 relies heavily on extensive trial-and-error, executing considerably more redundant or invalid actions to reach the same objectives.

\noindent\textbf{Real-world spatial complexity demands comprehensive evaluation.} Since no single model universally dominates—exemplified by 
% Qwen-3.5-397B-A17B excelling in Work tasks (35.6\%) but lagging behind Gemini-3.1-Pro in Travel and Digital domains
GPT-5 and Qwen-3.5-397B-A17B tying in Work\&Study tasks (16.9\%), GPT-5 leading Travel (6.8\%), and Gemini-3.1-Pro leading Digital domains (39.0\%)—our multifaceted benchmark is essential to accurately capture the diverse spatial capabilities required in real-world scenarios.

\subsection{Analysis}
\vspace{-0.2cm}
% As the optimal settings for inference-time factors—such as temperature, history window size, and motion parameterization—vary across different models.
% In addition to the aforementioned ablations, we report complementary slices of the main results. The indoor--outdoor split, complexity-mode split, and game-family split decompose the benchmark along different semantic and operational axes. 
% Detailed analyses of the multi-agent social split, image resolution, and camera field of view are provided in Appendix~\ref{sec:additional_benchmark_details}.
% Beyond aggregate metrics, we dissect the benchmark along multiple complementary axes to expose capability-specific bottlenecks. Specifically, we analyze the indoor--outdoor domain split, task complexity modes, multi-agent social coordination, and game-family breakdown, each targeting a distinct facet of spatial intelligence. We further probe perceptual factors (resolution, field of view) and inference-time hyperparameters (temperature, history window, action parameterization) 
Beyond aggregate metrics, we dissect the benchmark along complementary axes—indoor--outdoor split, task complexity, multi-agent coordination, and game-family breakdown—to expose capability-specific bottlenecks, and probe perceptual factors (resolution, field of view) and inference-time hyperparameters (temperature, history window, action parameterization).

\begin{figure*}[t]
% \vspace{-0.5cm}
\centering
\includegraphics[width=0.94\linewidth]{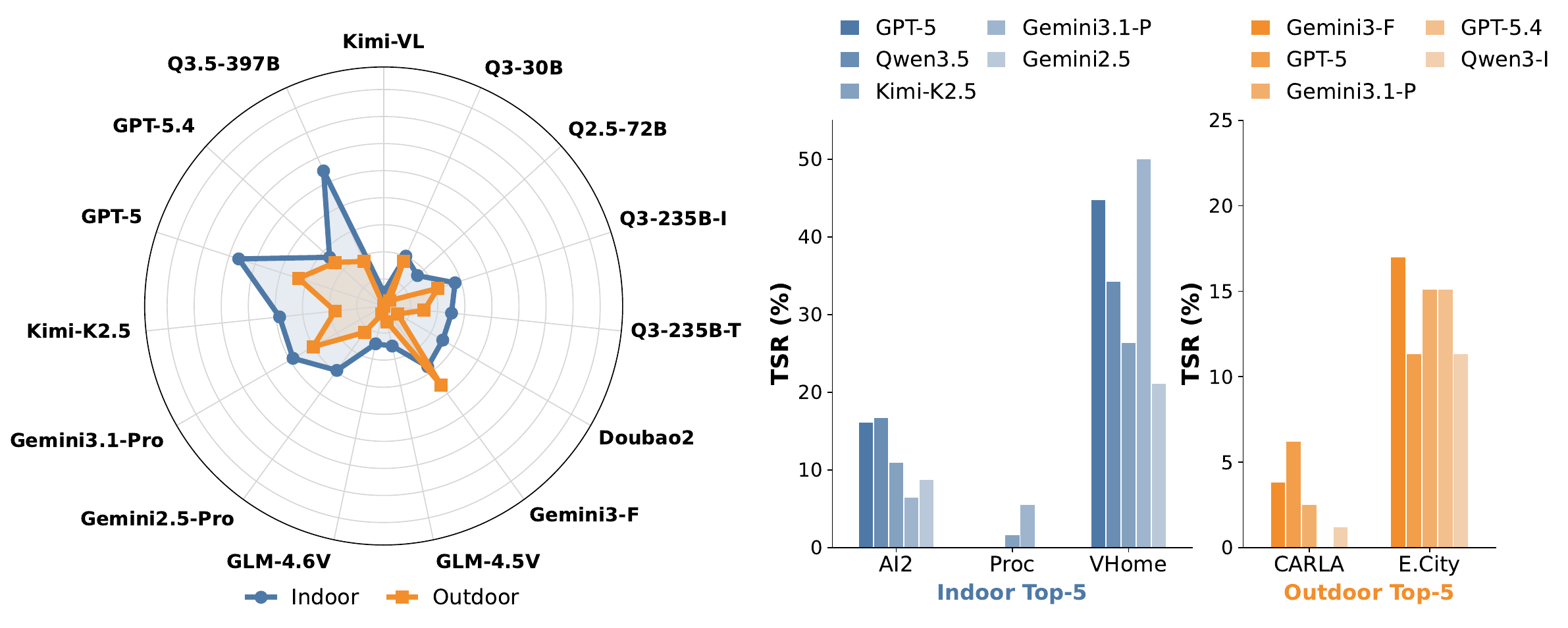}
\vspace{-0.1cm}
\caption{\textbf{Indoor and outdoor physical domains.} Overall TSR across indoor and outdoor physical environments, with environment-level bars for the top-five models in each domain.}
\label{fig:main_indoor_outdoor_tsr}
% \vspace{-0.3cm}
\end{figure*}

%jingyi: 这个图化成雷达图，两层，横坐标是模型，然后infoor、outdoor；右边是两列的图，从上往下indoor、outdoor最前的几个模型
%这个表格放附录
\textbf{Indoor vs. Outdoor Physical Environments.}
% We partition the single-agent physical benchmark into indoor environments (AI2THOR, ProcTHOR, and VirtualHome) and outdoor environments (CARLA and EmbodiedCity). We exclude both the suite of digital 3D games and the multi-agent environments from this comparison, as the latter primarily evaluate collaboration rather than scene comprehension. Fig.~\ref{fig:main_indoor_outdoor_tsr} reports the pooled domain-level TSR using a radar plot and further decomposes the top-five models in each domain across the constituent environments with grouped bar charts. Table~\ref{tab:main_indoor_outdoor_tsr} presents the corresponding environment-level TSR values. The overall columns reveal a significant domain shift: GPT-5 (22.5\%) and Qwen-3.5-397B-A17B (22.3\%) demonstrate the highest performance in indoor settings, whereas Gemini-3.1-Pro (38.3\%) and Gemini-3-Flash (32.3\%) achieve the highest results in outdoor settings. This performance divergence suggests distinct algorithmic biases among the models. The superior indoor performance of GPT-5 implies stronger capabilities in fine-grained object grounding and the low-level control of interactions. Conversely, the success of the Gemini series in outdoor environments indicates an advantage in long-horizon spatial reasoning and macro-level navigational planning.
We partition the single-agent physical benchmark into indoor and outdoor domains, excluding games and multi-agent setups to isolate scene comprehension. Fig.~\ref{fig:main_indoor_outdoor_tsr} and Table~\ref{tab:main_indoor_outdoor_tsr} (Appendix~\ref{sec:app_indoor_outdoor}) reveal a pronounced domain shift: 
% GPT-5 (23.9\%) and Qwen-3.5-397B-A17B (22.3\%) lead indoors, while Gemini-3.1-Pro (38.3\%) and Gemini-3-Flash (32.3\%) dominate outdoors. 
GPT-5 (14.1\%) and Qwen-3.5-397B-A17B (13.7\%) lead indoors, while Gemini-3-Flash (9.0\%) and GPT-5 (8.3\%) lead outdoors.
This divergence exposes distinct algorithmic biases. GPT-5's indoor superiority suggests robust fine-grained object grounding and low-level control, whereas the Gemini series' outdoor success highlights strengths in long-horizon spatial reasoning and macro-level navigation.

\begin{table*}[t]
\centering
\small
% \caption{\textbf{Efficiency Evaluation.} Main-benchmark SE across task categories. Bold and underlined entries denote the best and second-best values in each column, respectively. Models are grouped by open-source and closed-source families. Physical Overall is the average of Daily, Work, Entertain., Travel, and Social SE values; \texttt{--} denotes categories with no successful valid trajectory.}
\caption{\textbf{Efficiency Evaluation.} Main-benchmark SE across task categories. Bold and underlined denote the best and second-best per column. Physical Overall is the weighted mean over successful valid physical trajectories; \texttt{--} indicates no successful trajectory.}
\vspace{-0.1cm}
\label{tab:main_se}
\renewcommand{\arraystretch}{0.9}
\resizebox{0.96\textwidth}{!}{%
\begin{tabular}{lccccccc}
\toprule
\multirow{2}{*}{\textbf{Model}} & \multicolumn{6}{c}{\textbf{Physical}} & \multicolumn{1}{c}{\textbf{Digital}} \\
\cmidrule(lr){2-7}\cmidrule(lr){8-8}
 & \textbf{Daily} & \textbf{Work} & \textbf{Entertain.} & \textbf{Travel} & \textbf{Social} & \textbf{Overall} & \textbf{Entertain.} \\
\midrule
\rowcolor{purple!15} \multicolumn{8}{c}{\textbf{\textit{(A) Open-Source Models}}} \\
\midrule
\raisebox{-0.2ex}{\includegraphics[height=1em]{logos/qwen-color.pdf}}~Qwen2.5-VL-72B~\citep{bai2025qwen25vltechnicalreport} & 0.692 & 0.675 & 0.576 & \textbf{0.889} & 0.089 & 0.659 & \underline{0.757} \\
\raisebox{-0.2ex}{\includegraphics[height=1em]{logos/qwen-color.pdf}}~Qwen3-VL-30B-A3B~\citep{yang2025qwen3technicalreport} & 0.912 & \textbf{1.000} & \underline{0.889} & \underline{0.875} & 0.107 & \textbf{0.866} & 0.665 \\
\raisebox{-0.2ex}{\includegraphics[height=1em]{logos/qwen-color.pdf}}~Qwen3-VL-235B-Instruct~\citep{yang2025qwen3technicalreport} & 0.871 & \underline{0.864} & 0.863 & 0.482 & 0.145 & 0.737 & \textbf{0.889} \\
\raisebox{-0.2ex}{\includegraphics[height=1em]{logos/qwen-color.pdf}}~Qwen3-VL-235B-Thinking~\citep{yang2025qwen3technicalreport} & 0.732 & 0.683 & 0.681 & 0.574 & 0.137 & 0.625 & 0.666 \\
\raisebox{-0.2ex}{\includegraphics[height=1em]{logos/qwen-color.pdf}}~Qwen-3.5-397B-A17B~\citep{qwen35blog} & 0.689 & 0.605 & 0.594 & 0.690 & 0.172 & 0.609 & 0.694 \\
\raisebox{-0.2ex}{\includegraphics[height=1em]{logos/glm-color.pdf}}~GLM-4.5V~\citep{hong2025glm} & 0.841 & \textbf{1.000} & 0.707 & 0.550 & 0.164 & 0.659 & 0.634 \\
\raisebox{-0.2ex}{\includegraphics[height=1em]{logos/glm-color.pdf}}~GLM-4.6V~\citep{Glm4p6v} & \textbf{1.000} & 0.484 & 0.833 & 0.583 & \texttt{--} & 0.840 & 0.711 \\
\raisebox{-0.2ex}{\includegraphics[height=1em]{logos/kimi.pdf}}~Kimi-VL-A3B~\citep{team2025kimi} & 0.886 & 0.500 & \texttt{--} & \texttt{--} & \texttt{--} & 0.758 & 0.633 \\
\raisebox{-0.2ex}{\includegraphics[height=1em]{logos/kimi.pdf}}~Kimi-K2.5~\citep{team2026kimi} & 0.624 & 0.846 & 0.686 & 0.653 & 0.141 & 0.584 & 0.531 \\

\midrule
\rowcolor{blue!15} \multicolumn{8}{c}{\textbf{\textit{(B) Closed-Source Models}}} \\
\midrule
\raisebox{-0.2ex}{\includegraphics[height=1em]{logos/gemini-color.pdf}}~Gemini-2.5-Pro~\citep{comanici2025gemini25pushingfrontier} & 0.761 & 0.796 & \textbf{1.000} & 0.550 & \underline{0.236} & 0.688 & 0.647 \\
\raisebox{-0.2ex}{\includegraphics[height=1em]{logos/gemini-color.pdf}}~Gemini-3-Flash~\citep{gemini3flash} & 0.719 & 0.573 & 0.526 & 0.674 & 0.108 & 0.654 & 0.640 \\
\raisebox{-0.2ex}{\includegraphics[height=1em]{logos/gemini-color.pdf}}~Gemini-3.1-Pro~\citep{gemini3} & 0.814 & 0.665 & 0.541 & 0.788 & 0.175 & 0.736 & 0.717 \\
\raisebox{-0.2ex}{\includegraphics[height=1em]{logos/openai.pdf}}~GPT-5~\citep{singh2025openai} & 0.707 & 0.664 & 0.547 & 0.615 & 0.146 & 0.587 & 0.536 \\
\raisebox{-0.2ex}{\includegraphics[height=1em]{logos/openai.pdf}}~GPT-5.4~\citep{gpt5p4} & 0.796 & \textbf{1.000} & 0.726 & 0.647 & 0.200 & 0.745 & 0.702 \\
\raisebox{-0.2ex}{\includegraphics[height=1em]{logos/doubao-color.pdf}}~Doubao-2.0-Lite~\citep{seed} & \underline{0.954} & \textbf{1.000} & 0.875 & 0.854 & \textbf{0.326} & \underline{0.841} & 0.598 \\
\bottomrule
\end{tabular}%
}
\vspace{-0.4cm}
\end{table*}

\begin{figure*}[t]
% \vspace{-0.1cm}
\centering
\begin{subfigure}{0.321\textwidth}
\centering
\includegraphics[width=0.98\linewidth]{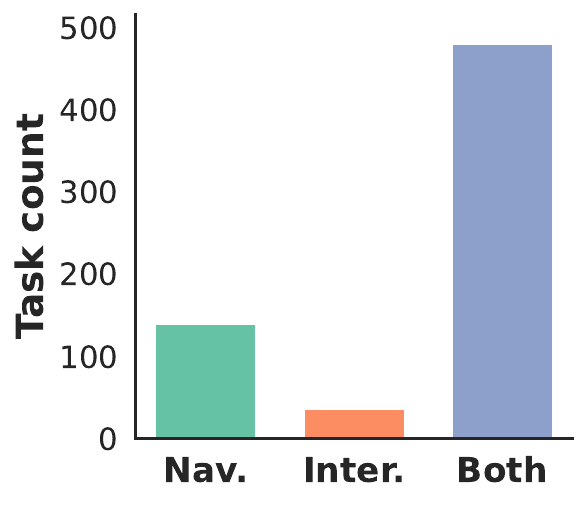}
\vspace{-0.25cm}
\caption{Task distribution.}
\end{subfigure}\hfill
\begin{subfigure}{0.321\textwidth}
\centering
\includegraphics[width=0.98\linewidth]{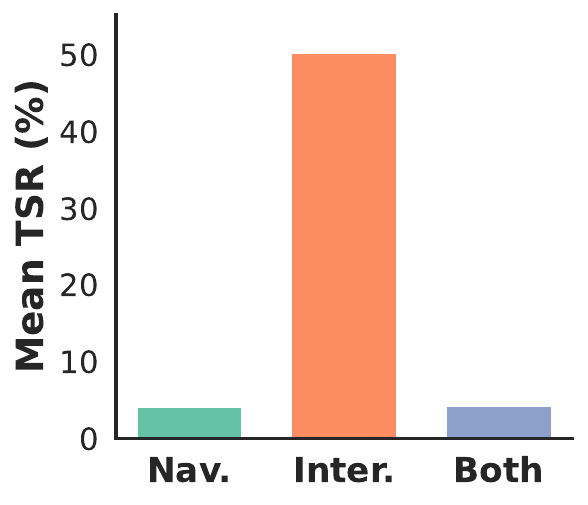}
\vspace{-0.25cm}
\caption{Performance.}
\end{subfigure}\hfill
\begin{subfigure}{0.321\textwidth}
\centering
\includegraphics[width=0.98\linewidth]{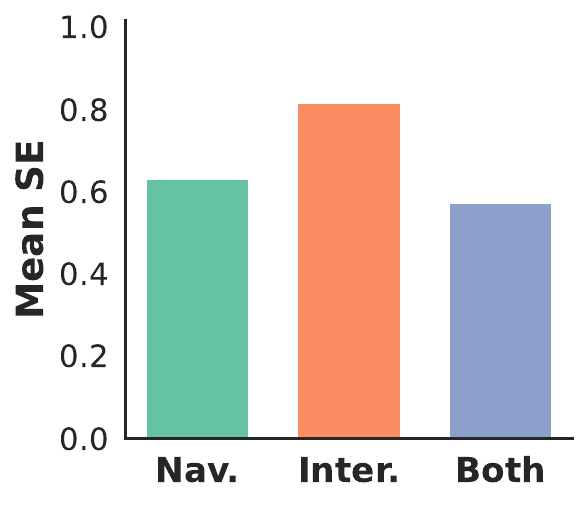}
\vspace{-0.25cm}
\caption{Efficiency.}
\end{subfigure}
% \vspace{-0.1cm}
\caption{\textbf{Complexity profile.} Task counts, mean TSR, and mean SE across the three parallel complexity modes in the physical benchmark.}
\label{fig:main_complexity_profile}
\vspace{-0.05cm}
\end{figure*}

\textbf{Complexity Modes.}
% We further categorize the physical benchmark by the action signatures introduced in Sec.~\ref{sec:taxonomy}, excluding the digital game suite from this analysis. Because the raw metadata lacks reliability for this split, we infer the complexity mode from the normalized golden action primitives: movement and viewpoint primitives define Navigation, object-state primitives define Interaction, and trajectories requiring both are assigned to Navigation--Interaction. Fig.~\ref{fig:main_complexity_profile} illustrates the distribution and outcomes of these tasks. The mean TSRs are 18.6\% for Navigation, 33.1\% for Interaction, and 9.9\% for Navigation--Interaction. This severe performance degradation in the combined mode demonstrates a compound bottleneck: executing precise object manipulations while maintaining long-term spatial progress presents a fundamentally greater challenge than performing either sub-task in isolation. Furthermore, the variance in leading models across modes---with Gemini-3.1-Pro leading Navigation (34.7\%) and GPT-5 dominating both Interaction (52.0\%) and Navigation--Interaction (23.3\%)---validates the design of the taxonomy. It confirms that these modes evaluate orthogonal capabilities rather than merely representing a singular scale of difficulty.
Categorizing tasks by the action signatures from Section~\ref{sec:taxonomy} reveals distinct complexity modes derived from golden action primitives: Navigation (movement and viewpoint), Interaction (object-state), and Navigation--Interaction (both). Fig.~\ref{fig:main_complexity_profile} illustrates a compound bottleneck: executing precise manipulations alongside long-term spatial progress (Navigation--Interaction, 4.2\% mean TSR) is demonstrably harder than Interaction (50.2\%). The variation in leading models across modes---with Gemini-series leading Navigation (8.6\%) and GPT-5 dominating Interaction (69.4\%) and the combined mode (12.1\%)---validates the taxonomy. These modes successfully evaluate orthogonal capabilities rather than a singular difficulty scale.

\begin{figure*}[t]
\vspace{-0.1cm}
\centering
\begin{minipage}[c]{0.55\textwidth}
\centering
% 关键：height 设小 + keepaspectratio=false 强制压扁
\includegraphics[width=0.96\linewidth,height=0.252\textheight,keepaspectratio=false]{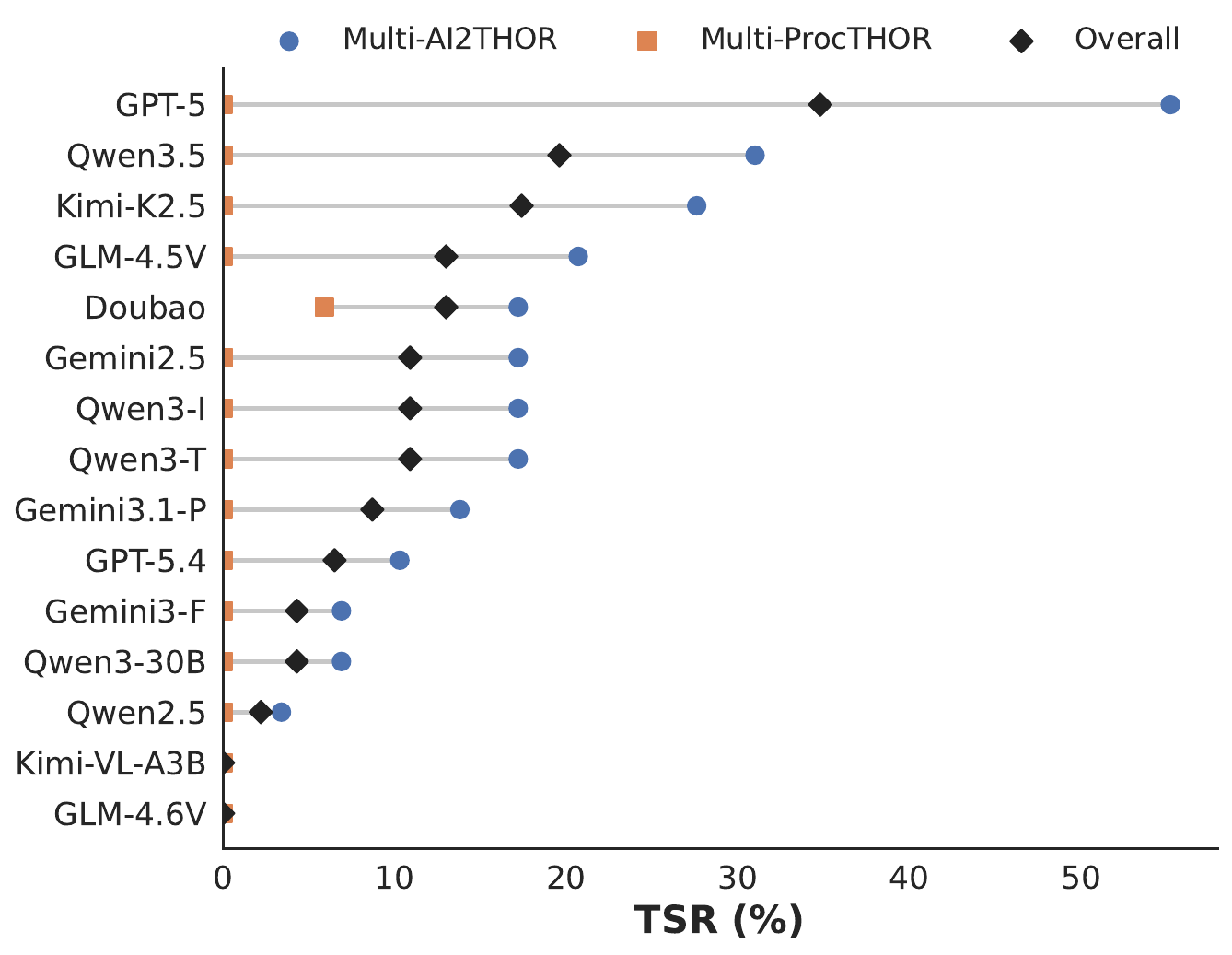}
\subcaption{Social collaboration TSR on Multi-AI2THOR and Multi-ProcTHOR; the dark marker denotes the pooled social score.}
\label{fig:main_multi_social_profile}
\end{minipage}\hfill
\begin{minipage}[c]{0.41\textwidth}
\centering
\begin{subfigure}{\linewidth}
\centering
\includegraphics[width=0.93\linewidth,height=0.132\textheight,keepaspectratio=false]{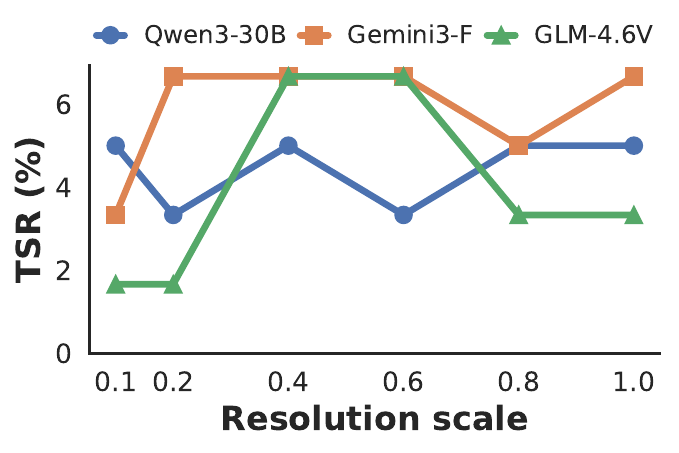}
\vspace{-0.22cm}
\caption{Resolution-scale probe on AI2THOR.}
\label{fig:observation_resolution}
\end{subfigure}

\vspace{-0.1cm}

\begin{subfigure}{\linewidth}
\centering
\includegraphics[width=0.94\linewidth,height=0.128\textheight,keepaspectratio=false]{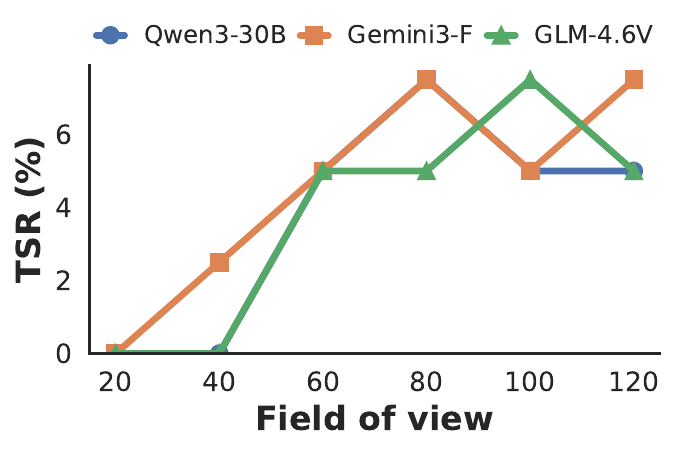}
\vspace{-0.2cm}
\caption{Camera-field-of-view probe on AI2THOR.}
\label{fig:observation_fov}
\end{subfigure}
\end{minipage}
\vspace{-0.1cm}
\caption{\textbf{Social and perceptual profiles.} Three complementary additional observations: multi-agent social performance, image-resolution sensitivity, and field-of-view sensitivity.}
\label{fig:social_observation_profiles}
\vspace{-0.2cm}
\end{figure*}

% \textbf{Multi-Agent Social Environments.}
% The Social Collaboration column in Table~\ref{tab:main} pools Multi-AI2THOR and Multi-ProcTHOR, but these environments stress different forms of coordination. Fig.~\ref{fig:main_multi_social_profile} therefore reports the two multi-agent environments separately. GPT-5 achieves the best pooled social TSR at 39.1\%, followed by Qwen-3.5-397B-A17B at 32.6\% and Qwen3-VL-235B-Thinking at 21.7\%. Most of this signal comes from Multi-AI2THOR, where object-centric cooperative routines are more frequently solved; Multi-ProcTHOR remains substantially harder, with the best models reaching only 5.9\%. The result suggests that current agents can sometimes coordinate in familiar, hand-authored indoor layouts, but procedural multi-agent layouts sharply reduce the reliability of shared progress tracking and role assignment.
 \textbf{Multi-Agent Social Environments.}
The Social Collaboration column in Table~\ref{tab:main} pools Multi-AI2THOR and Multi-ProcTHOR, but these environments stress different coordination patterns. Fig.~\ref{fig:main_multi_social_profile} reports them separately. 
% GPT-5 achieves the best pooled social TSR at 39.1\%, followed by Qwen-3.5-397B-A17B at 32.6\% and Qwen3-VL-235B-Thinking at 21.7\%
GPT-5 achieves the best pooled social TSR at 34.8\%, followed by Qwen-3.5-397B-A17B at 19.6\% and Kimi-K2.5 at 17.4\%. Most of this signal comes from Multi-AI2THOR, where object-centric cooperative routines are more frequently solved; Multi-ProcTHOR remains substantially harder, with the best models reaching only 5.9\%. This suggests that current agents can coordinate in familiar, hand-authored indoor layouts, but procedural multi-agent layouts sharply reduce the reliability of shared progress tracking and role assignment.

% \begin{figure*}[t]
% \vspace{-0.4cm}
% \centering
% % 1. 继续增大左侧的宽度占比，使左图等比放大后更高
% \begin{minipage}[c]{0.56\textwidth}
% \centering
% \includegraphics[width=0.96\linewidth,height=0.27\textheight]{img/main_multi_social_profile.pdf}
% \subcaption{Social collaboration TSR on Multi-AI2THOR and Multi-ProcTHOR; the dark marker denotes the pooled social score.}
% \label{fig:main_multi_social_profile}
% \end{minipage}\hfill
% % 2. 继续减小右侧宽度
% \begin{minipage}[c]{0.41\textwidth}
% \centering
% \begin{subfigure}{\linewidth}
% \centering
% \includegraphics[width=\linewidth]{img/ablation_resolution_tsr_polished.pdf}
% % \vspace{0.2em}
% \vspace{-0.63cm}
% \caption{Resolution-scale probe on AI2THOR.}
% \label{fig:observation_resolution}
% \end{subfigure}

% % 3. �� 加大右侧上下两张图的间距 (从0.7em加大到1.5em或2em)
% % 这样可以撑起右侧整体排版的高度，使其与变高后的左图在视觉上更对齐平衡

% \begin{subfigure}{\linewidth}
% \centering
% \includegraphics[width=\linewidth]{img/ablation_fov_tsr_polished.pdf}
% \vspace{-0.63cm}
% \caption{Camera-field-of-view probe on AI2THOR.}
% \label{fig:observation_fov}
% \end{subfigure}
% \end{minipage}
% \vspace{-0.18cm}
% \caption{\textbf{Social and perceptual profiles.} Three complementary additional observations: multi-agent social performance, image-resolution sensitivity, and field-of-view sensitivity.}
% \label{fig:social_observation_profiles}
% \vspace{-0.6cm}
% \end{figure*}

\begin{figure*}[t]
\vspace{-0.2cm}
\centering
\begin{subfigure}{0.322\textwidth}
\centering
\includegraphics[width=\linewidth]{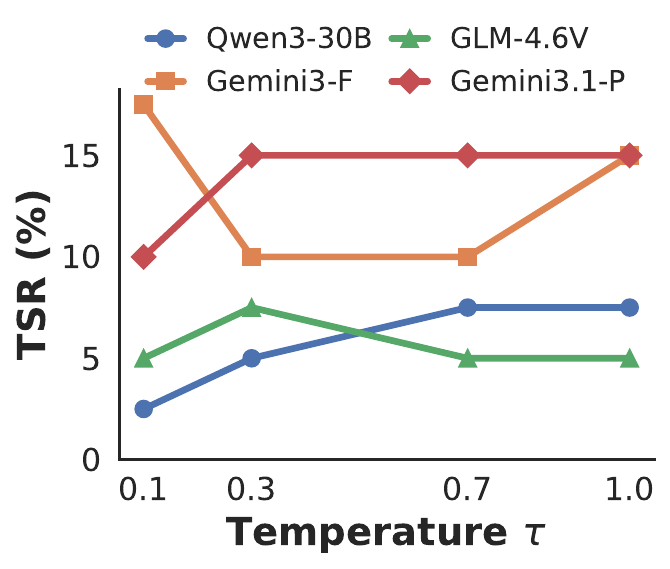}
\vspace{-0.4cm}
\caption{Temperature}
\end{subfigure}\hfill
\begin{subfigure}{0.322\textwidth}
\centering
\includegraphics[width=\linewidth]{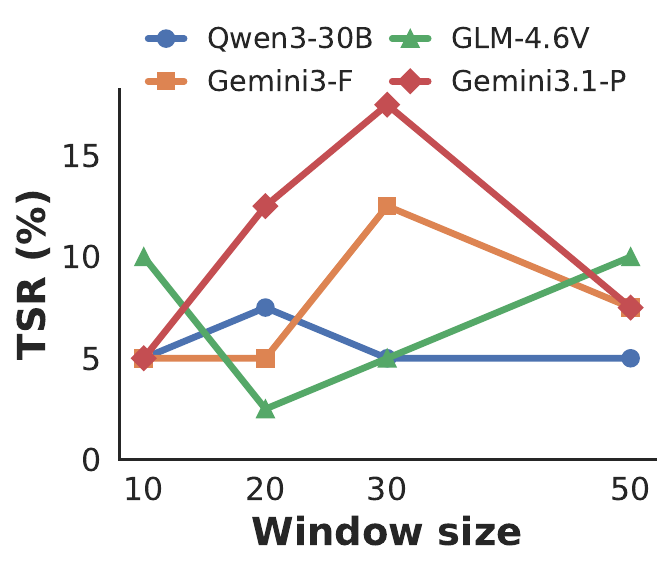}
\vspace{-0.4cm}
\caption{History window}
\end{subfigure}
\vspace{0.5em}
\begin{subfigure}{0.322\textwidth}
\centering
\includegraphics[width=\linewidth]{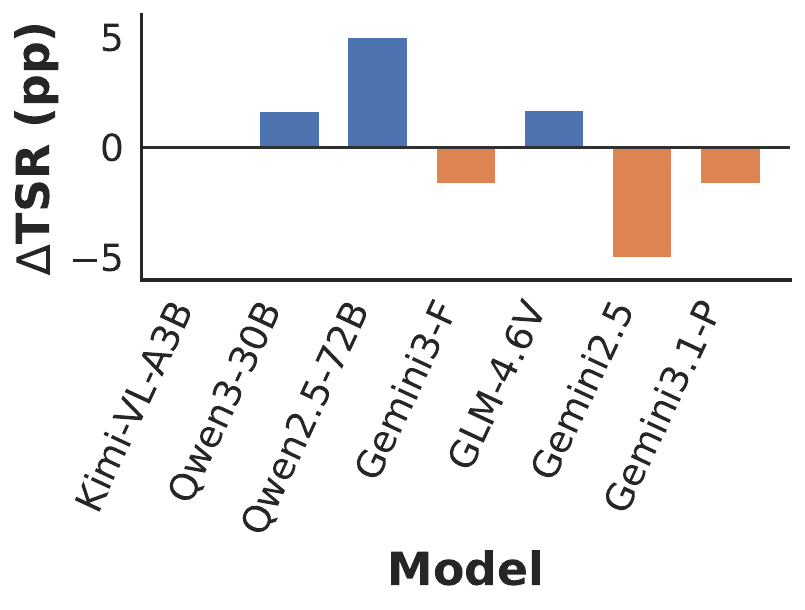}
\vspace{-0.4cm}
\caption{Continuous--discrete TSR gap}
\end{subfigure}
\vspace{-0.22cm}
\caption{\textbf{Ablation trends.} TSR under temperature and history-window settings, together with the signed TSR gap between continuous and discrete action parameterizations.}
\label{fig:ablation_combined}
\vspace{-0.2cm}
\vspace{-0.2cm}
\end{figure*}

\textbf{Game-Level Breakdown.}
We further analyze performance across five digital game families (see Table~\ref{tab:main_game_tsr} in Appendix~\ref{sec:app_game_breakdown}). The results reveal that while top models achieve strong reactive control in navigation and Snake tasks, they systematically struggle with games requiring explicit geometric reasoning and multi-step state transformations (e.g., Rubik and Block3D), indicating that spatial manipulation remains a fundamental bottleneck for multimodal agents.

\textbf{Perceptual Factors.} 
% Figures~\ref{fig:observation_resolution} and \ref{fig:observation_fov} indicate that enriching the visual evidence—either through higher image resolutions or wider fields of view—generally improves task success rates. Higher resolutions preserve finer object details and local geometry, while wider views expose more scene content per step, thereby mitigating perceptual myopia. Consequently, higher-resolution inputs are preferable when computational budgets permit. However, regarding the field of view, despite the performance advantages of wider coverage, we set the default to 60 to more closely approximate a human-like first-person viewing condition.
Figs.~\ref{fig:observation_resolution} and \ref{fig:observation_fov} show that perceptual configuration affects the visual evidence available to the agent. The resolution curve remains comparatively flat and locally non-monotonic, suggesting that image resolution does not materially affect interactive spatial understanding (see Appendix~\ref{sec:app_observation_sensitivity} for the visualization). 
%Once the relevant scene structure is visible, increasing pixel density adds limited geometric information; the dominant failures instead arise from belief updating, route planning, and action selection.
For field of view, higher-FOV settings outperform narrow views overall, but gains plateau and vary by setting. We nevertheless keep the default field of view at 60 to more closely approximate a human-like first-person viewing condition.

\textbf{Inference-time Hyperparameters.} 
% We also ablate temperature, history window size, and action parameterization on a subset in Fig.~\ref{fig:ablation_combined}: temperature induces only a 1.3\% TSR spread a cross models, optimal window size is model-dependent, and continuous versus discrete motion shows no universal winner. Results show limited sensitivity across three factors. Detailed analysis are in Appendix~\ref{sec:additional_benchmark_details}.  
We also ablate temperature, history window size, and action parameterization on a subset in Fig.~\ref{fig:ablation_combined}: temperature sensitivity is marginal, and optimal choices for window size and motion type are completely model-dependent. As no single setting proves universally optimal, we default to standard configurations. Detailed analysis is provided in Appendix~\ref{sec:additional_benchmark_details}.

% \vspace{-0.2cm}\subsection{Analysis}

% \textbf{Failure Mode Breakdown.}
% We manually inspect 100 failed trajectories and categorize failures into: (i)~\textit{Spatial Disorientation}---the agent loses track of its position and cannot return to a target location; (ii)~\textit{Object Hallucination}---the agent issues manipulation actions on objects not present in the current view; (iii)~\textit{Premature Termination}---the agent issues \texttt{DONE} or \texttt{FAIL} before the goal is satisfied; and (iv)~\textit{Action Loop}---the agent cycles through the same sequence of ineffective actions until the step budget is exhausted.
% % TODO: insert failure-mode pie chart or bar chart.
% Spatial disorientation and action loops together account for over 60\% of all failure cases. We select one representative bad case for detailed analysis in the following section.

% \textbf{Correlation Between Spatial Reasoning and TSR.}
% We correlate each model's TSR on \textsc{SpatialWorld} with its score on established spatial VQA benchmarks (SpatialScore~\cite{wu2025spatialscore}, EmbSpatial~\cite{du2024embspatial}).
% A moderate positive correlation ($r \approx 0.6$) is observed, suggesting that static spatial reasoning ability is a necessary but insufficient predictor of interactive task-solving performance.

% \textbf{Human Performance.}
% Human annotators achieve XX\% TSR with an average SE of XX, providing a strong upper bound and demonstrating that \textsc{SpatialWorld} tasks are solvable with human-level spatial cognition.

\section{Related Work}
% \vspace{-0.1cm}
% \vspace{-0.2cm}
\subsection{Multimodal Agents}
Multimodal agents are characterized by unified perception and state representations over multimodal inputs (\textit{e.g.,} text, images, and videos), together with multi-step planning and decision-making to complete complex tasks via tool use or direct actions~\cite{xie2024large,zhang2024mm,li2025system,bai2025qwen25vltechnicalreport,team2023gemini}. Early research followed two main directions: one focused on improving multimodal foundation models for stronger representation and understanding, establishing generalizable perceptual features and vision-language alignment~\cite{team2025kimi,liu2024llava,guo2025seed1}. The other developed agentic interaction and long-horizon task execution frameworks, enabling iterative planning and sequential decision-making in interactive environments~\cite{hong2024cogagent,he2024webvoyager,wang2023drivemlm}.
The multimodal agents have been applied across a wide range of settings, including image understanding and editing~\cite{zheng2025deepeyes,zhang2025thyme,fan2024videoagent}, computer use via screen-based interaction~\cite{wang2025ui,wang2024mobile,wang2025opencua}, and physical embodied environments~\cite{driess2023palm,lin2024vila,elnoor2025vlm,chen2025robogpt}.

\vspace{-0.2cm}\subsection{3D Environment Simulators}
A wide range of 3D simulation platforms have been developed to support spatial reasoning, navigation, and autonomous decision-making across diverse domains~\cite{li2021igibson,xia2018gibson,ramakrishnan2021habitat,xiang2020sapien,yu2020meta}.
For indoor embodied interaction, AI2-THOR~\cite{kolve2017ai2} provides interactive, near-photorealistic scenes with rich object affordances for studying task-oriented manipulation.
Habitat~\cite{savva2019habitat} offers an efficient modular framework with configurable sensors and agent embodiments, and is widely used for navigation, instruction following, and embodied question answering.
For autonomous driving, CARLA~\cite{dosovitskiy2017carla} and MetaDrive~\cite{li2022metadrive} simulate urban traffic with flexible sensor suites and dynamic actors, serving as standard testbeds for learning-based perception and control.
Recent efforts further expand simulator coverage and realism~\cite{fan2022minedojo,bu2025agibot,yang2025drivearena,lan2025autobio}, broadening the space of scenarios and world dynamics available for evaluation.
Despite these advances, existing platforms and their associated scenarios remain domain-specific (\textit{e.g.,} indoor manipulation vs.\ urban driving) and adopt heterogeneous task definitions and interfaces, making it difficult to compare general, open-ended task-solving ability across settings.

\vspace{-0.2cm}\subsection{Spatial Reasoning of Multimodal Agents}
Spatial reasoning in multimodal agents refers to the ability to ground goals in real-world perceptual observations and maintain an evolving spatial belief about the environment under observability~\cite{zhu2024llava,cai2025scaling,xu2024pointllm,chen2024spatialvlm}, enabling agents to localize objects, infer relative motion relationships, and support reliable planning and action in physical space~\cite{liu2025spatial,zhou2025robotracer,liu2025spatialcot}.
Spatial reasoning has been primarily evaluated through visual question answering benchmarks~\cite{johnson2017clevr,hudson2019gqa,liu2023visual,azuma2022scanqa,ma2022sqa3d,du2024embspatial,wu2025spatialscore,duan2024vlmevalkit} under fixed 2D observations, with recent evaluations extending to 3D and video settings~\cite{yang2025mmsi,lin2025ost,hong20233d,yang2025thinking} to test whether models can build and recall spatial structure from sequential observations.
Multi-turn interaction is essential for spatial reasoning, as agents must make sequential decisions to gather information and update spatial beliefs over time. However, most existing multi-step benchmarks are either grounded in 2D screen~\cite{xie2024osworld,rawles2024androidworld,koh2024visualwebarena} or adopt heterogeneous embodied interfaces and observation assumptions~\cite{liu2025spatial,li2023m3dbench,gholami2025spatial}.
For instance, EmbodiedBench~\cite{yang2025embodiedbench} evaluates agents that process visual and sensor data to predict low-level atomic navigation or manipulation actions.
% By comparison, we focus on evaluating foundation multimodal agents under a unified setting that uses only egocentric visual observations and executes high-level instruction actions, such as turning left.
In contrast, \textsc{SpatialWorld} evaluates foundation multimodal agents under a unified closed-loop protocol across heterogeneous 3D environments. 
Agents receive only egocentric RGB observations and high-level task instructions, and their success is determined by task-specific terminal-state verifiers rather than static answer matching or low-level action prediction accuracy.

\vspace{-0.3cm}\section{Conclusion}\vspace{-0.2cm}

We introduced \textsc{SpatialWorld}, a unified benchmark designed to evaluate the interactive spatial reasoning of MLLMs. By abstracting simulator-specific complexities into a shared text-based interface, our benchmark rigorously assesses an agent's capacity for active egocentric exploration and decision-making under partial observability. Extensive evaluations of 15 leading MLLMs reveal a critical vulnerability: while current models excel at static scene perception, they struggle profoundly with dynamic physical environments—exhibiting low task success rates, severe execution inefficiencies, and high domain variance. These bottlenecks underscore a fundamental gap in robust interactive spatial reasoning and long-horizon planning. We envision \textsc{SpatialWorld} as a foundational testbed to shift the paradigm of MLLM research from passive observation to the realization of general-purpose spatial agents capable of seamless interaction in the real world.

\newpage

% \begin{ack}
% Use unnumbered first level headings for the acknowledgments. All acknowledgments

% Do {\bf not} include this section in the anonymized submission, only in the final paper. You can use the \texttt{ack} environment provided in the style file to automatically hide this section in the anonymized submission.
% \end{ack}

\bibliography{1_ref}
\bibliographystyle{plainnat}

%%%%%%%%%%%%%%%%%%%%%%%%%%%%%%%%%%%%%%%%%%%%%%%%%%%%%%%%%%%%%%%%%%%%%%%%%%%%%%%
% APPENDIX
%%%%%%%%%%%%%%%%%%%%%%%%%%%%%%%%%%%%%%%%%%%%%%%%%%%%%%%%%%%%%%%%%%%%%%%%%%%%%%%
\clearpage
\appendix
% \vspace{-0.1cm}

\section{Additional Benchmark Details}
\label{sec:additional_benchmark_details}

This section provides supplementary details for the benchmark construction and evaluation described in Sections~\ref{sec:eval} and the main benchmark protocol. We include a full benchmark comparison table, task-specific evaluation criteria, and fine-grained performance breakdowns across indoor/outdoor environments and digital game families.

\subsection{Detailed Benchmark Comparison}
\label{sec:app_benchmark_compare}

Table~\ref{tab:benchmark_compare_full} presents an extended comparison of \textsc{SpatialWorld} against existing spatial reasoning benchmarks spanning three major categories: ImageQA, VideoQA, and embodied-agent evaluation. We compare along five critical dimensions: (1) whether the benchmark provides a unified cross-platform interface that abstracts away environment-specific APIs, (2) whether agents interact with a dynamic, interactive environment rather than answering questions over static inputs, (3) whether observations are captured from a first-person (egocentric) perspective, (4) whether the input modality is purely visual without auxiliary structured data such as depth maps or object coordinates, and (5) whether the output is expressed in natural language form. As shown in the table, existing ImageQA and VideoQA benchmarks~\citep{wang2024spatialeval,ma20253dsrbench,wu2025spatialscore,xu2025spatialbench,yang2025thinking} predominantly evaluate passive spatial understanding through static question answering, lacking interactive environments and unified interfaces. Embodied benchmarks~\citep{shridhar2020alfred,zhang2024vlabench,cheng2025embodiedeval,yang2025embodiedbench} introduce interactivity but typically sacrifice one or more desirable properties---either requiring privileged non-visual inputs, lacking language-form outputs, or being restricted to a single simulation platform. In contrast, \textsc{SpatialWorld} is the only benchmark that simultaneously satisfies all five criteria, enabling a holistic evaluation of active spatial reasoning under realistic embodied constraints across diverse environments.

\begin{table*}[htbp]
\centering
\scriptsize
\caption{\textbf{Detailed spatial benchmark comparison.} Extended version of Table~\ref{tab:benchmark_compare}, including the full set of representative ImageQA, VideoQA, and embodied-agent benchmarks used to motivate the benchmark construction.}
\label{tab:benchmark_compare_full}
\renewcommand{\arraystretch}{1.2}
\resizebox{\textwidth}{!}{%
\begin{tabular}{llcccccc}
\toprule
\textbf{Type} & \textbf{Benchmark} & \textbf{Instances} & \makecell[c]{\textbf{Unified cross-}\\\textbf{platform interface}} & \makecell[c]{\textbf{Interactive}\\\textbf{env.}} & \makecell[c]{\textbf{First-person}\\\textbf{observation}} & \makecell[c]{\textbf{Vision-only}\\\textbf{input}} & \makecell[c]{\textbf{Language-form}\\\textbf{output}} \\
\midrule
\multirow{7}{*}{ImageQA}
& SpatialEval~\citep{wang2024spatialeval} & 4635 & \xmark & \xmark & \xmark & \cmark & \cmark \\
& 3DSRBench~\citep{ma20253dsrbench} & 2772 & \xmark & \xmark & \xmark & \cmark & \cmark \\
& EmbSpatial-Bench~\citep{du2024embspatial} & 3640 & \xmark & \xmark & \cmark & \cmark & \cmark \\
& SpatialScore~\citep{wu2025spatialscore} & 5025 & \xmark & \xmark & \xmark & \cmark & \cmark \\
& OmniSpatial~\citep{jia2025omnispatial} & 8400 & \xmark & \xmark & \xmark & \cmark & \cmark \\
& EASI~\citep{cai2025holistic} & 24k & \xmark & \xmark & \xmark & \cmark & \cmark \\
& MMSI-Bench~\citep{yang2025mmsi} & 1000 & \xmark & \xmark & \xmark & \cmark & \cmark \\
\midrule
\multirow{5}{*}{VideoQA}
& SpatialBench~\citep{xu2025spatialbench} & 3193 & \xmark & \xmark & \cmark & \cmark & \cmark \\
& SITE~\citep{wang2025site} & 8068 & \xmark & \xmark & \xmark & \cmark & \cmark \\
& MMSI-Video-Bench~\citep{lin2025mmsi} & 1106 & \xmark & \xmark & \xmark & \cmark & \cmark \\
& VSI-Bench~\citep{yang2025thinking} & 5130 & \xmark & \xmark & \cmark & \cmark & \cmark \\
& OST-Bench~\citep{lin2025ost} & 10k & \xmark & \xmark & \cmark & \cmark & \cmark \\
\midrule
\multirow{6}{*}{Embodied Bench}
& ALFRED~\citep{shridhar2020alfred} & 25.7k & \xmark & \cmark & \cmark & \cmark & \xmark \\
& VLABench~\citep{zhang2024vlabench} & 100 & \xmark & \cmark & \cmark & \xmark & \xmark \\
& EAI~\citep{li2024eai} & 438 & \cmark & \cmark & \xmark & \xmark & \cmark \\
& EmbodiedEval~\citep{cheng2025embodiedeval} & 328 & \xmark & \cmark & \cmark & \cmark & \xmark \\
& EmbodiedBench~\citep{yang2025embodiedbench} & 1128 & \cmark & \cmark & \cmark & \xmark & \cmark \\
& EmbodiedCity~\citep{gao2024embodiedcity} & 87.1k & \xmark & \cmark & \cmark & \cmark & \cmark \\
\midrule
\textbf{Ours}
& \textsc{SpatialWorld} & 760 & \cmark & \cmark & \cmark & \cmark & \cmark \\
\bottomrule
\end{tabular}%
}
\end{table*}

\subsection{Task-Specific Evaluation Details}
\label{sec:eval_details}

While \textsc{SpatialWorld} primarily evaluates tasks using the binary TSR based on exact goal satisfaction, certain environments require task-specific adaptations. 

For instance, in the Snake3D environment, exact completion is too sparse to separate weak partial progress from complete failure. Therefore, instead of using a binary success indicator, we evaluate performance by reporting a scale-normalized discrete score. This is calculated by dividing the achieved snake score by the spatial edge length of the game environment, providing a more granular measure of the agent's progress.

\subsection{Indoor vs.\ Outdoor Performance Breakdown}
\label{sec:app_indoor_outdoor}

Table~\ref{tab:main_indoor_outdoor_tsr} reports the per-environment TSR for all 15 evaluated models, partitioned into indoor (AI2THOR, ProcTHOR, VirtualHome) and outdoor (CARLA, EmbodiedCity) domains. Multi-agent environments are excluded here and analyzed separately in Fig.~\ref{fig:main_multi_social_profile}. The overall columns pool the environments within each domain. This fine-grained breakdown reveals that GPT-5 and Qwen-3.5-397B-A17B dominate in indoor scenarios requiring precise object grounding, whereas GPT-5 and Gemini-3-Flash lead in outdoor scenarios that demand long-range navigation and spatial planning.

\begin{table*}[htbp]
\centering
\small
\caption{\textbf{Indoor-outdoor.} The TSR (\%) of the single-agent physical benchmark across indoor and outdoor environments. Bold and underlined entries denote the best and second-best values in each column, respectively. Multi-agent environments are excluded here and analyzed separately in Figure~\ref{fig:main_multi_social_profile}. The overall columns pool the environments within each domain and are located at the right edge of each domain group.}
\label{tab:main_indoor_outdoor_tsr}
\renewcommand{\arraystretch}{0.9}
\resizebox{0.98\textwidth}{!}{%
\begin{tabular}{lccccccc}
\toprule
\multirow{2}{*}{\textbf{Model}} & \multicolumn{4}{c}{\textbf{Indoor}} & \multicolumn{3}{c}{\textbf{Outdoor}} \\
\cmidrule(lr){2-5}\cmidrule(lr){6-8}
 & \textbf{AI2THOR} & \textbf{ProcTHOR} & \textbf{VHome} & \textbf{Overall} & \textbf{CARLA} & \textbf{E.City} & \textbf{Overall} \\
\midrule
\rowcolor{purple!15} \multicolumn{8}{c}{\textbf{\textit{(A) Open-Source Models}}} \\
\midrule
\raisebox{-0.2ex}{\includegraphics[height=1em]{logos/qwen-color.pdf}}~Qwen2.5-VL-72B~\citep{bai2025qwen25vltechnicalreport} & 5.1 & 0.0 & 10.5 & 4.2 & 1.2 & 0.0 & 0.8 \\
\raisebox{-0.2ex}{\includegraphics[height=1em]{logos/qwen-color.pdf}}~Qwen3-VL-30B-A3B~\citep{yang2025qwen3technicalreport} & 6.4 & 0.0 & 10.5 & 5.0 & 1.2 & 9.4 & 4.5 \\
\raisebox{-0.2ex}{\includegraphics[height=1em]{logos/qwen-color.pdf}}~Qwen3-VL-235B-Instruct~\citep{yang2025qwen3technicalreport} & 7.7 & 0.8 & 21.1 & 6.9 & 1.2 & 11.3 & 5.3 \\
\raisebox{-0.2ex}{\includegraphics[height=1em]{logos/qwen-color.pdf}}~Qwen3-VL-235B-Thinking~\citep{yang2025qwen3technicalreport} & 9.0 & 0.0 & 5.3 & 6.3 & 1.2 & 7.5 & 3.8 \\
\raisebox{-0.2ex}{\includegraphics[height=1em]{logos/qwen-color.pdf}}~Qwen-3.5-397B-A17B~\citep{qwen35blog} & \textbf{16.7} & 0.0 & 34.2 & \underline{13.7} & 2.5 & 7.5 & 4.5 \\
\raisebox{-0.2ex}{\includegraphics[height=1em]{logos/glm-color.pdf}}~GLM-4.5V~\citep{hong2025glm} & 4.5 & 0.0 & 10.5 & 3.8 & 1.2 & 1.9 & 1.5 \\
\raisebox{-0.2ex}{\includegraphics[height=1em]{logos/glm-color.pdf}}~GLM-4.6V~\citep{Glm4p6v} & 4.8 & 0.0 & 5.3 & 3.6 & 0.0 & 1.9 & 0.8 \\
\raisebox{-0.2ex}{\includegraphics[height=1em]{logos/kimi.pdf}}~Kimi-VL-A3B~\citep{team2025kimi} & 1.6 & 0.0 & 2.6 & 1.3 & 0.0 & 0.0 & 0.0 \\
\raisebox{-0.2ex}{\includegraphics[height=1em]{logos/kimi.pdf}}~Kimi-K2.5~\citep{team2026kimi} & 10.9 & \underline{1.6} & 26.3 & 9.7 & \underline{3.8} & 5.7 & 4.5 \\

\midrule
\rowcolor{blue!15} \multicolumn{8}{c}{\textbf{\textit{(B) Closed-Source Models}}} \\
\midrule
\raisebox{-0.2ex}{\includegraphics[height=1em]{logos/gemini-color.pdf}}~Gemini-2.5-Pro~\citep{comanici2025gemini25pushingfrontier} & 8.7 & 0.0 & 21.1 & 7.4 & 1.2 & 5.7 & 3.0 \\
\raisebox{-0.2ex}{\includegraphics[height=1em]{logos/gemini-color.pdf}}~Gemini-3-Flash~\citep{gemini3flash} & 7.7 & 0.8 & 21.1 & 6.9 & \underline{3.8} & \textbf{17.0} & \textbf{9.0} \\
\raisebox{-0.2ex}{\includegraphics[height=1em]{logos/gemini-color.pdf}}~Gemini-3.1-Pro~\citep{gemini3} & 6.4 & \textbf{5.5} & \textbf{50.0} & 9.7 & 2.5 & \underline{15.1} & 7.5 \\
\raisebox{-0.2ex}{\includegraphics[height=1em]{logos/openai.pdf}}~GPT-5~\citep{singh2025openai} & \underline{16.1} & 0.0 & \underline{44.7} & \textbf{14.1} & \textbf{6.2} & 11.3 & \underline{8.3} \\
\raisebox{-0.2ex}{\includegraphics[height=1em]{logos/openai.pdf}}~GPT-5.4~\citep{gpt5p4} & 8.4 & 0.0 & 15.8 & 6.7 & 0.0 & \underline{15.1} & 6.0 \\
\raisebox{-0.2ex}{\includegraphics[height=1em]{logos/doubao-color.pdf}}~Doubao-2.0-Lite~\citep{seed} & 6.1 & 0.0 & 28.9 & 6.3 & 1.2 & 1.9 & 1.5 \\
\bottomrule
\end{tabular}%
}
\end{table*}

\subsection{Game-Level Performance Breakdown}
\label{sec:app_game_breakdown}

Table~\ref{tab:main_game_tsr} presents the per-game-family TSR for all evaluated models across five digital game environments: Block3D (B3D), Maze, Maze3D (M3D), Rubik's Cube, and Snake. Each column pools the available levels for the corresponding game. The Snake environment normalizes scores by the spatial edge length and caps each level contribution at 100\%.

Gemini-3.1-Pro demonstrates the highest overall efficacy (39.0\%), driven by strong results on Block3D (40.0\%) and Snake (90.0\%), while Gemini-3-Flash leads Rubik's Cube (50.0\%). In contrast, topological traversal tasks expose different architectural strengths: Qwen3-VL-235B-Thinking excels in both 2D pathfinding (Maze, 70.0\%) and 3D perspective navigation (Maze3D, 32.0\%), whereas GPT-5 is strongest on Snake (91.2\%). This performance divergence reveals that while top-tier architectures demonstrate robust proficiency in reactive visual-motor alignment and topological routing, they systematically falter on tasks demanding explicit geometric reasoning and complex structural state transformations. The generally low success rates on Rubik and Block3D emphasize that multi-step spatial manipulation remains a fundamental bottleneck for embodied intelligence.

\begin{table*}[htbp]
\centering
\small
\caption{\textbf{Performance of Game.} TSR (\%) by game families. Bold and underlined entries denote the best/second-best values in each column. B3D denotes Block3D, and M3D denotes Maze3D.}
\label{tab:main_game_tsr}
\renewcommand{\arraystretch}{0.9}
\resizebox{0.9\textwidth}{!}{%
\begin{tabular}{lcccccc}
\toprule
\textbf{Model} & \textbf{B3D} & \textbf{Maze} & \textbf{M3D} & \textbf{Rubik} & \textbf{Snake} & \textbf{Overall} \\
\midrule
\rowcolor{purple!15} \multicolumn{7}{c}{\textbf{\textit{(A) Open-Source Models}}} \\
\midrule
\raisebox{-0.2ex}{\includegraphics[height=1em]{logos/qwen-color.pdf}}~Qwen2.5-VL-72B~\citep{bai2025qwen25vltechnicalreport} & 0.0 & 30.0 & 4.0 & 0.0 & 5.0 & 7.6 \\
\raisebox{-0.2ex}{\includegraphics[height=1em]{logos/qwen-color.pdf}}~Qwen3-VL-30B-A3B~\citep{yang2025qwen3technicalreport} & 5.0 & 25.0 & 0.0 & 5.0 & 6.2 & 7.9 \\
\raisebox{-0.2ex}{\includegraphics[height=1em]{logos/qwen-color.pdf}}~Qwen3-VL-235B-Instruct~\citep{yang2025qwen3technicalreport} & 10.0 & 5.0 & 4.0 & 5.0 & 1.2 & 5.0 \\
\raisebox{-0.2ex}{\includegraphics[height=1em]{logos/qwen-color.pdf}}~Qwen3-VL-235B-Thinking~\citep{yang2025qwen3technicalreport} & 5.0 & \textbf{70.0} & \textbf{32.0} & 10.0 & 23.8 & 28.3 \\
\raisebox{-0.2ex}{\includegraphics[height=1em]{logos/qwen-color.pdf}}~Qwen-3.5-397B-A17B~\citep{qwen35blog} & 5.0 & \underline{65.0} & 20.0 & 5.0 & 36.2 & 26.0 \\
\raisebox{-0.2ex}{\includegraphics[height=1em]{logos/glm-color.pdf}}~GLM-4.5V~\citep{hong2025glm} & 0.0 & 25.0 & 12.0 & 0.0 & 36.2 & 14.5 \\
\raisebox{-0.2ex}{\includegraphics[height=1em]{logos/glm-color.pdf}}~GLM-4.6V~\citep{Glm4p6v} & 0.0 & 30.0 & 4.0 & 5.0 & 2.5 & 8.1 \\
\raisebox{-0.2ex}{\includegraphics[height=1em]{logos/kimi.pdf}}~Kimi-VL-A3B~\citep{team2025kimi} & 0.0 & 0.0 & 8.0 & 5.0 & 2.5 & 3.3 \\
\raisebox{-0.2ex}{\includegraphics[height=1em]{logos/kimi.pdf}}~Kimi-K2.5~\citep{team2026kimi} & 5.0 & 40.0 & \underline{28.0} & 10.0 & 72.5 & 31.0 \\

\midrule
\rowcolor{blue!15} \multicolumn{7}{c}{\textbf{\textit{(B) Closed-Source Models}}} \\
\midrule
\raisebox{-0.2ex}{\includegraphics[height=1em]{logos/gemini-color.pdf}}~Gemini-2.5-Pro~\citep{comanici2025gemini25pushingfrontier} & 5.0 & 60.0 & 16.0 & 5.0 & 81.2 & 32.6 \\
\raisebox{-0.2ex}{\includegraphics[height=1em]{logos/gemini-color.pdf}}~Gemini-3-Flash~\citep{gemini3flash} & \underline{35.0} & 10.0 & 16.0 & \textbf{50.0} & 85.0 & \underline{38.1} \\
\raisebox{-0.2ex}{\includegraphics[height=1em]{logos/gemini-color.pdf}}~Gemini-3.1-Pro~\citep{gemini3} & \textbf{40.0} & 5.0 & 20.0 & \underline{45.0} & \underline{90.0} & \textbf{39.0} \\
\raisebox{-0.2ex}{\includegraphics[height=1em]{logos/openai.pdf}}~GPT-5~\citep{singh2025openai} & 0.0 & \underline{65.0} & 20.0 & 10.0 & \textbf{91.2} & 36.4 \\
\raisebox{-0.2ex}{\includegraphics[height=1em]{logos/openai.pdf}}~GPT-5.4~\citep{gpt5p4} & 5.0 & 15.0 & 24.0 & 0.0 & 12.5 & 11.9 \\
\raisebox{-0.2ex}{\includegraphics[height=1em]{logos/doubao-color.pdf}}~Doubao-2.0-Lite~\citep{seed} & 0.0 & 35.0 & 16.0 & 10.0 & 65.0 & 24.8 \\
\bottomrule
\end{tabular}%
}
\end{table*}

\vspace{-0.2cm}\section{Ablation Studies}
\label{sec:app_ablation}

This section provides detailed ablation results complementing the analysis in Section~\ref{sec:eval}. As summarized in Fig.~\ref{fig:ablation_combined} of the main text, we study three inference-time factors---temperature, history window size, and action parameterization---and find that their optimal settings are model-dependent rather than universal. Below we discuss each factor in detail.

\subsection{Temperature}
\label{sec:app_temperature}
% Fig.~\ref{fig:ablation_combined} demonstrates that the optimal temperature varies across model families, although the aggregate effect remains limited. The pooled TSR fluctuates by approximately 1.9 percentage points between the extreme temperature settings, with distinct models peaking at different values. Consequently, we avoid model-specific tuning of $\tau$. Following the evaluation protocol of OSWorld~\citep{xie2024osworld}, which utilizes $\tau=1.0$ alongside nucleus sampling, we set $\tau=1.0$ for all models. This configuration ensures protocol uniformity across model families while preserving moderate sampling diversity throughout long-horizon interactions.
Fig.~\ref{fig:ablation_combined} illustrates the impact of sampling temperature on performance. We observe that nearly all models, with the sole exception of Gemini-3-Flash, achieve their optimal performance at $\tau=1.0$. Consequently, following the evaluation protocol of OSWorld~\citep{xie2024osworld}, we set $\tau=1.0$ for all models. This configuration ensures protocol uniformity across model families while preserving moderate sampling diversity throughout long-horizon interactions.

\subsection{History Window Size}
\label{sec:app_history_window}
% Fig.~\ref{fig:ablation_combined} indicates that the optimal sliding window size also varies across models. Extended visual histories do not yield universal improvements: highly capable models benefit from larger context windows, whereas less capable models plateau or degrade when the retained history becomes excessively long. This pattern suggests that history length interacts with context utilization capacity rather than functioning as a straightforward scaling mechanism. 
% Therefore, the main benchmark uses $w=30$ as the default context window, and the ablation treats the corresponding main-run result under the same evaluation protocol as the $w=30$ baseline when a separate ablation run is absent.
Fig.~\ref{fig:ablation_combined} illustrates that $w=30$ serves as an optimal sliding window size across most evaluated models. We observe that while performance improves with initial increases in window size, it tends to plateau or slightly diminish beyond $w=30$. This suggests that a context window of 30 frames provides sufficient temporal information, and further extending the visual history does not yield universal gains. Consequently, we adopt $w=30$ as the default context window for the main benchmark. In our ablation analysis, the corresponding main-run result is treated as the $w=30$ baseline to ensure consistency.

% Therefore, the main benchmark employs the largest context consistently supported by the evaluated models. Because GPT-5.4 imposes the tightest input limit by restricting each turn to 50 images, we set $w=50$ throughout the default evaluation.

\subsection{Continuous versus Discrete Motion}
\label{sec:app_motion}
Fig.~\ref{fig:ablation_combined} reveals no universal preference for continuous over discrete motion. In the action-parameterization panel, each bar reports $\Delta\mathrm{TSR}=\mathrm{TSR}_{\mathrm{continuous}}-\mathrm{TSR}_{\mathrm{discrete}}$ in percentage points; positive values favor continuous action parameters, whereas negative values favor the discrete interface. Because the optimal action parameterization is model-dependent, we utilize discrete actions in the main benchmark. This decision maintains interface consistency across environments and avoids biasing the leaderboard toward a control granularity that favors a specific model family.

\section{Observation Sensitivity Analysis}
\label{sec:app_observation_sensitivity}
\begin{figure}
\vspace{-0.4cm}
    \centering
    \includegraphics[width=0.95\linewidth]{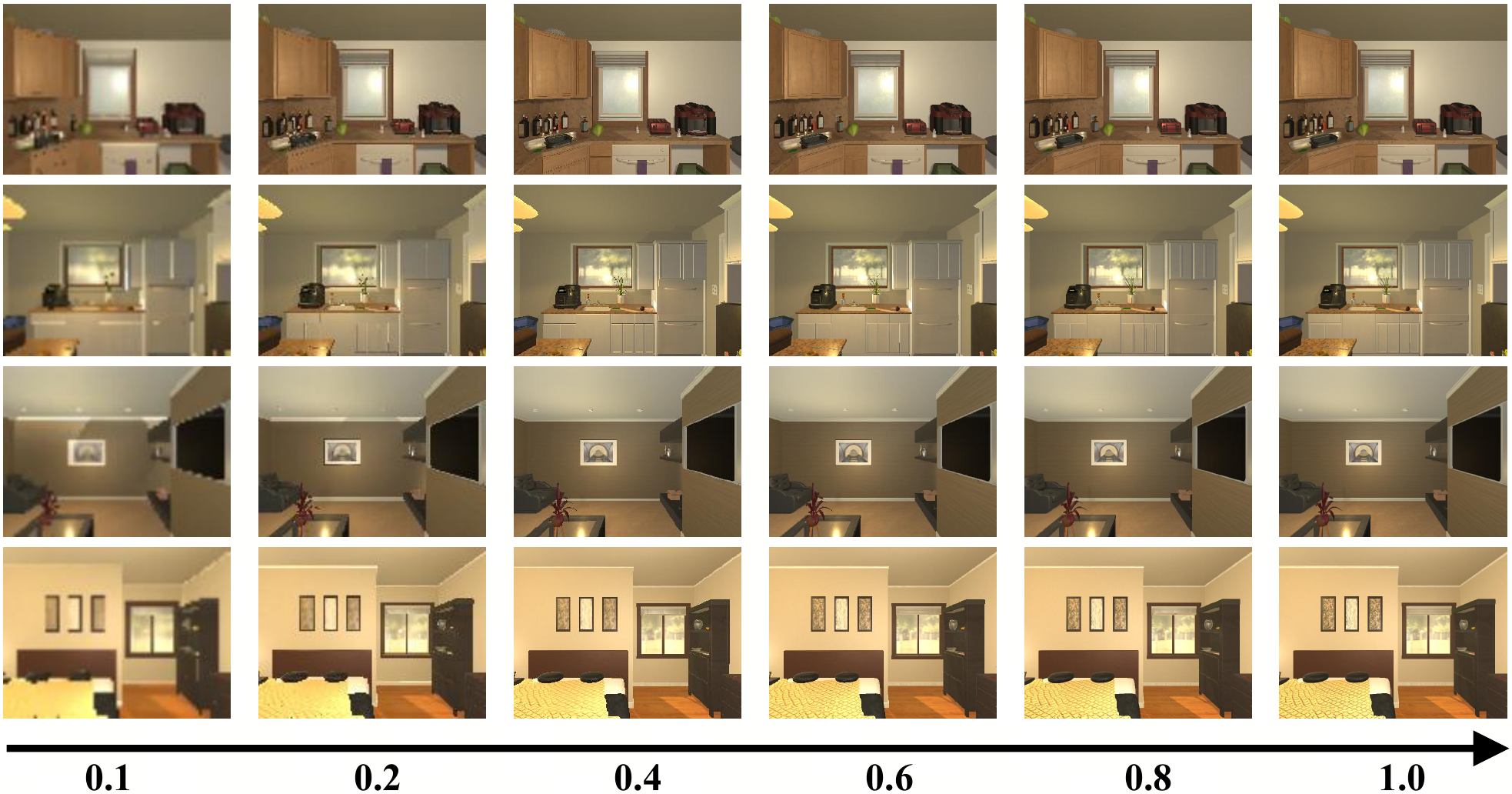}
    % \caption{\textbf{Data construction pipeline of SpatialWorld.} We adopt this pipeline across all environments, and have specifically trained annotators and hired expert teams to assist with annotation and verification. }
    \caption{\textbf{Observation Sensitivity Analysis under the Same Viewpoint with Varying Resolutions.} We progressively increase the resolution ratio along the x-axis, reaching the highest clarity at 1.0.} 
    % We provide example figures for all environments in Fig.~\ref{fig:placeholder} and include detailed environment introduction documents in the open-source code.
    \label{fig:fov}
    \vspace{-0.15cm}
\end{figure}

Fig.~\ref{fig:fov} shows visualizations at various resolutions. Decreasing resolution does not impair the model's spatial reasoning because spatial perception relies on physical, projective, and ray relationships, which are unaffected by lower resolutions. The visual results confirm that the model maintains accurate spatial awareness.

  % \subsection{Effect of Image Resolution}
  % \label{sec:app_resolution_sensitivity}

\vspace{0.2cm}\section{Environment Suite}
\label{sec:app_env_suite}

\textsc{SpatialWorld} uses its environment suite as the main source of domain diversity rather than as a passive collection of scenes.
We wrap eight 3D backends with a shared agent-side API, so agents interact through the same observation and action abstractions while the environments retain their native differences in scale, dynamics, object affordances, and scene generation.
This design exposes a broad spectrum of spatial demands under one evaluation surface: hand-authored indoor worlds test fine-grained object grounding, procedural houses test layout generalization, urban and aerial environments test long-range progress estimation, multi-agent variants test coordination, and digital games isolate abstract geometric reasoning.
As a result, cross-environment comparisons reflect genuine differences in 3D task-solving requirements rather than changes in the agent interface.
The suite is organized into three families.

\textbf{Indoor Simulation.}
AI2-THOR~\cite{kolve2017ai2} provides hand-designed, near-photorealistic indoor scenes with explicit object affordances, physical interactions, and state changes; it therefore anchors the benchmark in fine-grained object grounding and manipulation.
ProcTHOR~\cite{DBLP:conf/nips/DeitkeVHWESHKKM22} extends this setting through procedurally generated houses, increasing layout diversity and reducing reliance on a fixed set of manually authored rooms.
VirtualHome~\cite{DBLP:conf/cvpr/PuigRBLWF018} complements both environments by representing household activities as executable programs, making it suitable for scripted daily routines that require temporally ordered actions.
Building on the same indoor affordance space, the multi-agent variants of AI2-THOR and ProcTHOR introduce cooperative tasks in which success depends not only on locating or manipulating objects, but also on coordinating role-specific progress across agents.

\textbf{Outdoor Navigation.}
CARLA~\cite{dosovitskiy2017carla} shifts the benchmark from indoor object interaction to urban driving, where agents must reason over road topology, long-range route progress, and dynamic traffic context.
EmbodiedCity further broadens the outdoor setting to aerial city navigation, emphasizing landmark-based localization, altitude-aware movement, and macroscopic spatial planning over dense urban layouts.
Together, the outdoor environments test whether a model can maintain progress estimates and termination decisions when the relevant spatial evidence is distributed across a much larger field than a household room.

\textbf{Digital Game Environments.}
Indoor and outdoor simulators are indispensable for realistic embodied tasks, but they do not exhaust the space of 3D reasoning problems: their difficulty is often entangled with photorealistic semantics, simulator affordances, and natural scene priors.
We therefore add lightweight 3D games as controlled probes that isolate abstract spatial abilities under closed-loop interaction.
Block3D requires three-view geometric counting from orthographic projections; Maze and Maze3D test topological planning in two- and three-dimensional layouts; Snake3D stresses incremental state tracking under self-occlusion and limited free space; and Rubik's Cube evaluates spatial transformation and configuration reasoning.
These games broaden the benchmark beyond household and urban navigation by exposing spatial structures that are rare in realistic simulators but central to general 3D reasoning.

\vspace{-0.2cm}\section{Human Annotation}
\label{sec:app_human_annotation}

To ensure the quality and reproducibility of \textsc{SpatialWorld}, all 760 tasks undergo a rigorous three-stage human annotation process (see Fig.~\ref{fig:pipeline} for an overview of the pipeline). In the first stage, annotators design each task by specifying the natural-language instruction and configuring the initial environment state. In the second stage, annotators independently solve each task inside the simulator, recording the ground-truth terminal state and a reference action trajectory. In the third stage, annotators cross-check each other's work to verify task feasibility, instruction clarity, and evaluation-script correctness. Table~\ref{tab:evaluation_scripts} presents representative examples of the resulting annotated evaluation scripts, which retrieve dynamic state data (e.g., spatial coordinates, object containment, vehicle kinematics) from the 3D simulators to reliably assess functional correctness in open-ended physical environments.

\begin{table*}[htbp] 
\centering
\caption{Examples of our annotated execution-based evaluation scripts in SpatialWorld. The scripts retrieve dynamic state data (e.g., spatial coordinates, object containment, vehicle kinematics) from the 3D simulators to reliably assess functional correctness in open-ended physical environments.}
\label{tab:evaluation_scripts}

\renewcommand{\arraystretch}{1.0} 

\begin{tabularx}{\textwidth}{@{} 
    >{\centering\arraybackslash}m{0.22\textwidth} 
    >{\centering\arraybackslash}m{0.17\textwidth} 
    >{\raggedright\arraybackslash}m{0.22\textwidth} 
    >{\centering\arraybackslash}m{0.38\textwidth}
@{}} 
\toprule
\textbf{Overview State} & \textbf{Initial State} & \multicolumn{1}{c}{\textbf{Task Instruction}} & \multicolumn{1}{l}{\hspace*{1em}\hspace*{1em}\textbf{Success Condition}}  \\
\midrule

% --- Row 1: AI2-THOR ---
\adjustbox{valign=m}{\includegraphics[width=\linewidth, height=2.2cm]{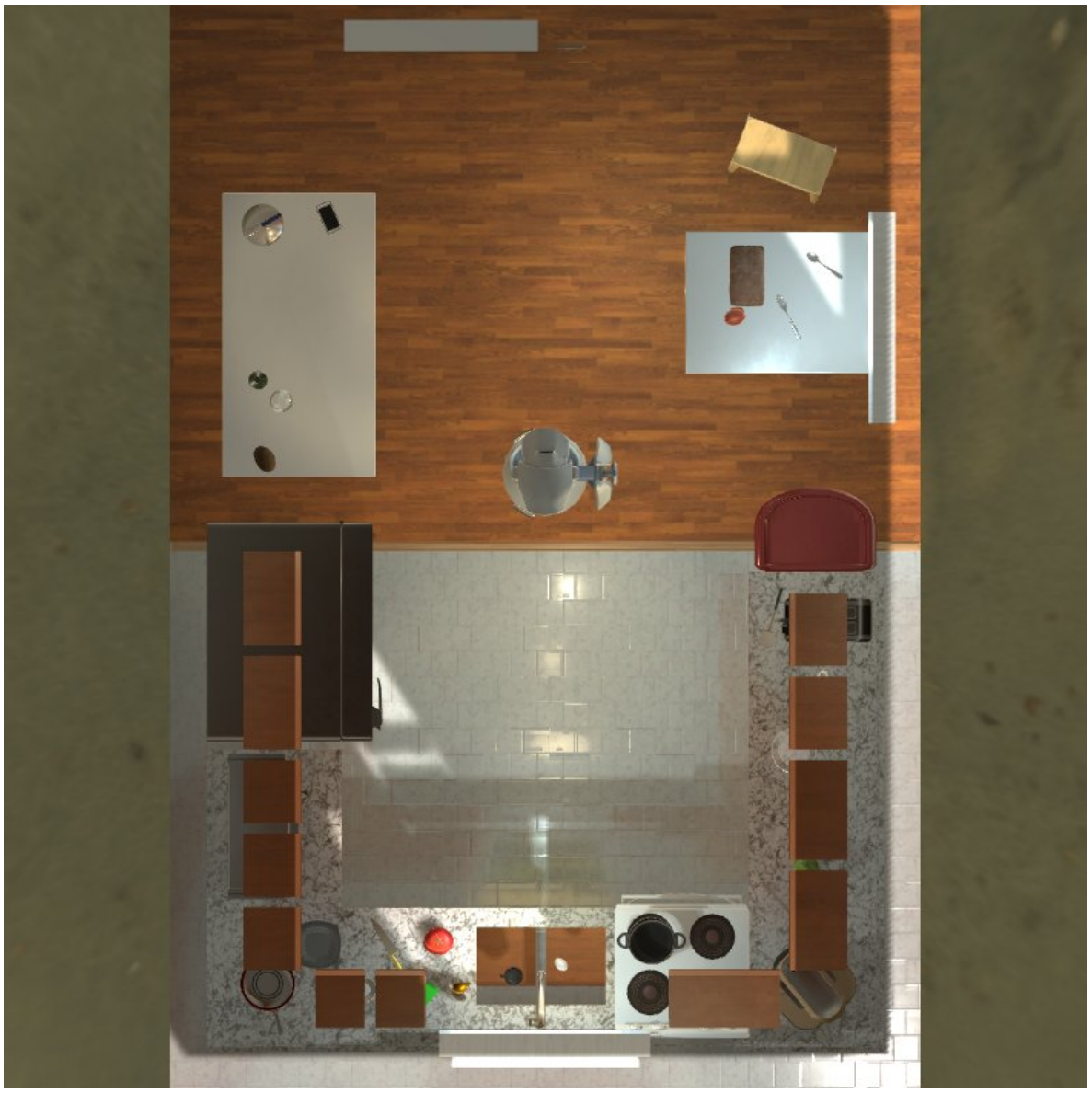}}
& 
\adjustbox{valign=m}{\includegraphics[width=\linewidth, height=2.2cm]{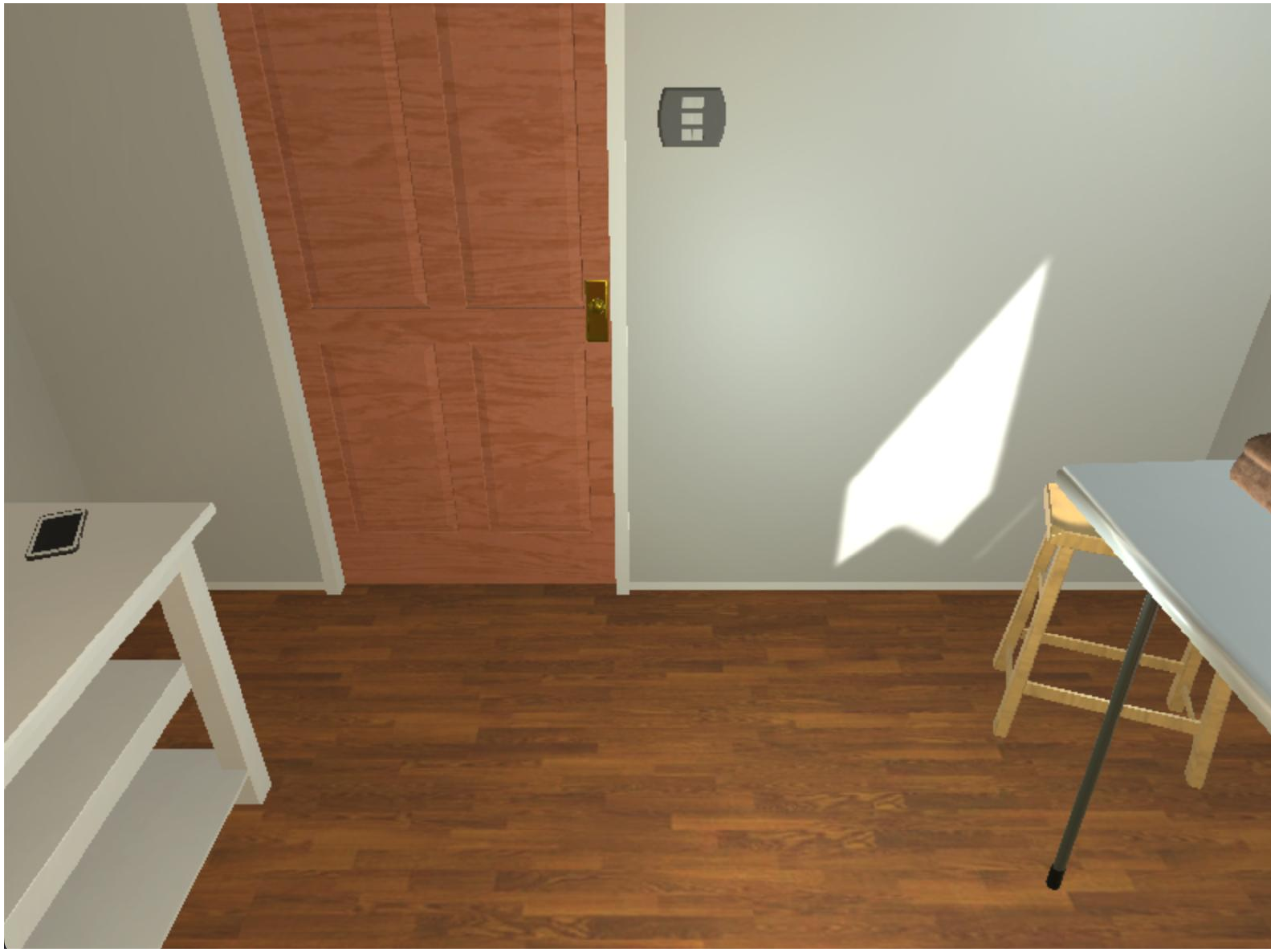}}
& 
\parbox[c]{\linewidth}{
\textit{I found the lettuce was rotten; please help me throw it in the trash.}
}
& 
\parbox[c]{\linewidth}{
\footnotesize
\begin{tabular}{@{}l@{}}
\hspace*{1em}\hspace*{1em}\textbf{object\_in\_receptacle} \\
\hspace*{1em}\hspace*{1em}\hspace*{1em} \textbf{object\_type}:\\
\hspace*{1em}\hspace*{1em}\hspace*{1em}\hspace*{1em}  Lettuce \\
\hspace*{1em}\hspace*{1em}\hspace*{1em} \textbf{receptacle\_type}: \\
\hspace*{1em}\hspace*{1em}\hspace*{1em}\hspace*{1em}  GarbageCan
\end{tabular}
} \\
\midrule

% --- Row 2: CARLA ---
\adjustbox{valign=m}{\includegraphics[width=\linewidth, height=2.2cm]{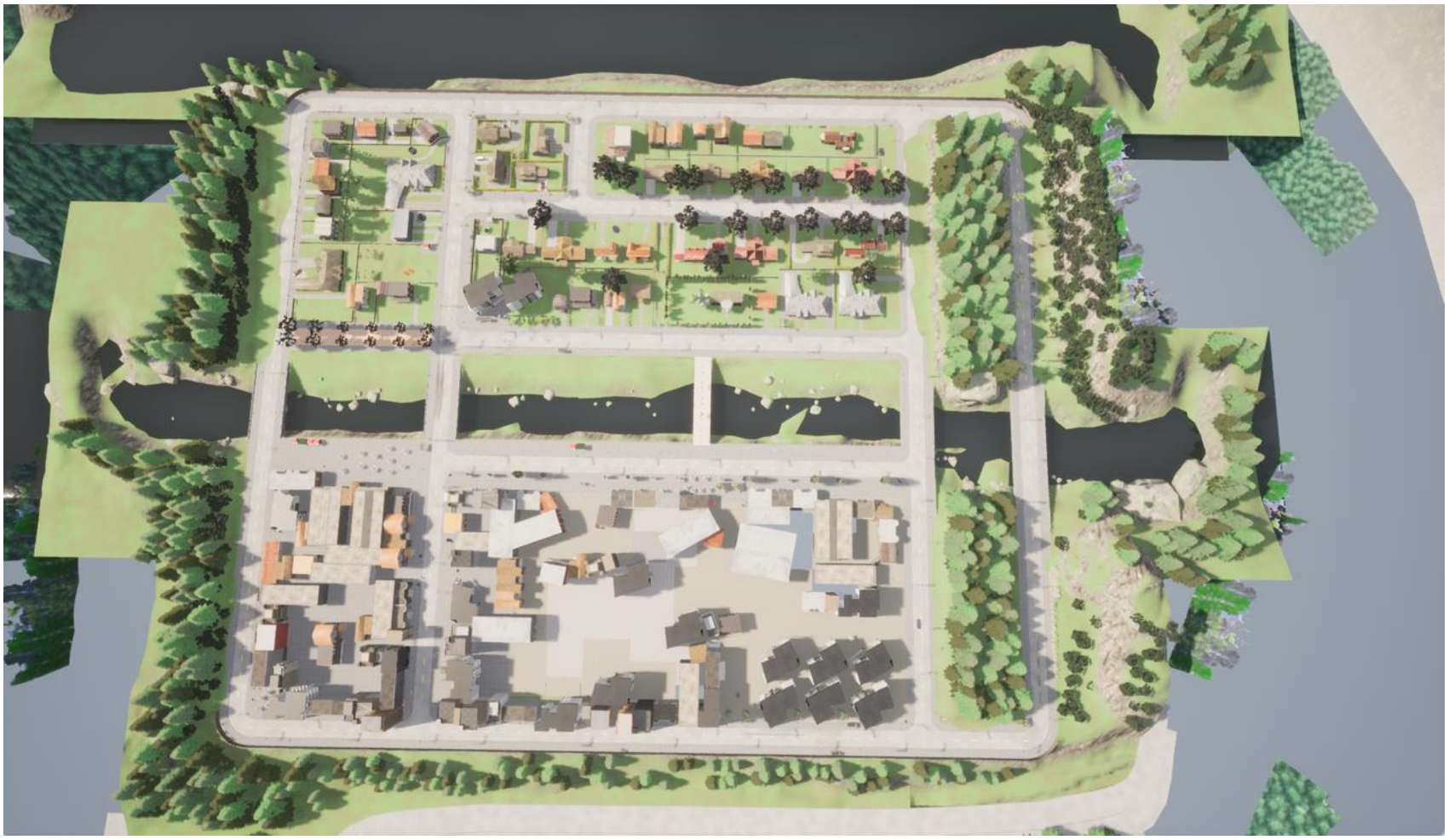}}
& 
\adjustbox{valign=m}{\includegraphics[width=\linewidth, height=2.2cm]{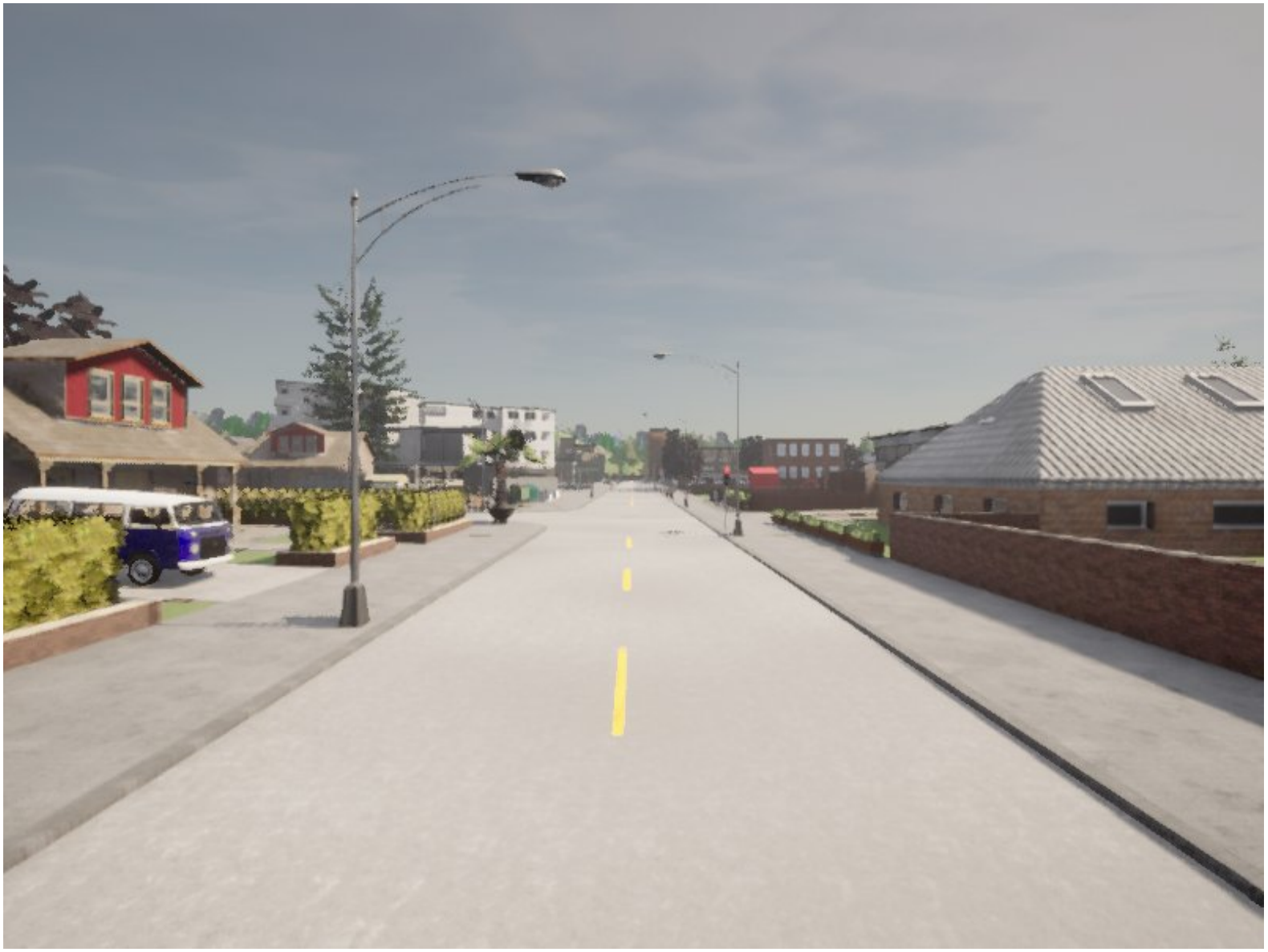}}
& 
\parbox[c]{\linewidth}{
\textit{Walk to the position marked by the red line in the screenshot. You can turn and move in any direction.}
}
& 
\parbox[c]{\linewidth}{
\footnotesize
\begin{tabular}{@{}l@{}}
\hspace*{1em}\hspace*{1em}\textbf{distance\_to\_waypoint} \\
\hspace*{1em}\hspace*{1em}\hspace*{1em} \textbf{target\_location}: \\
\hspace*{1em}\hspace*{1em}\hspace*{1em}\hspace*{1em}  [41.9, 32.9, 1.2] \\
\hspace*{1em}\hspace*{1em}\hspace*{1em} \textbf{threshold\_label}: \\
\hspace*{1em}\hspace*{1em}\hspace*{1em}\hspace*{1em}  medium
\end{tabular}
} \\
\midrule

% --- Row 3: VirtualHome ---
\adjustbox{valign=m}{\includegraphics[width=\linewidth, height=2.2cm]{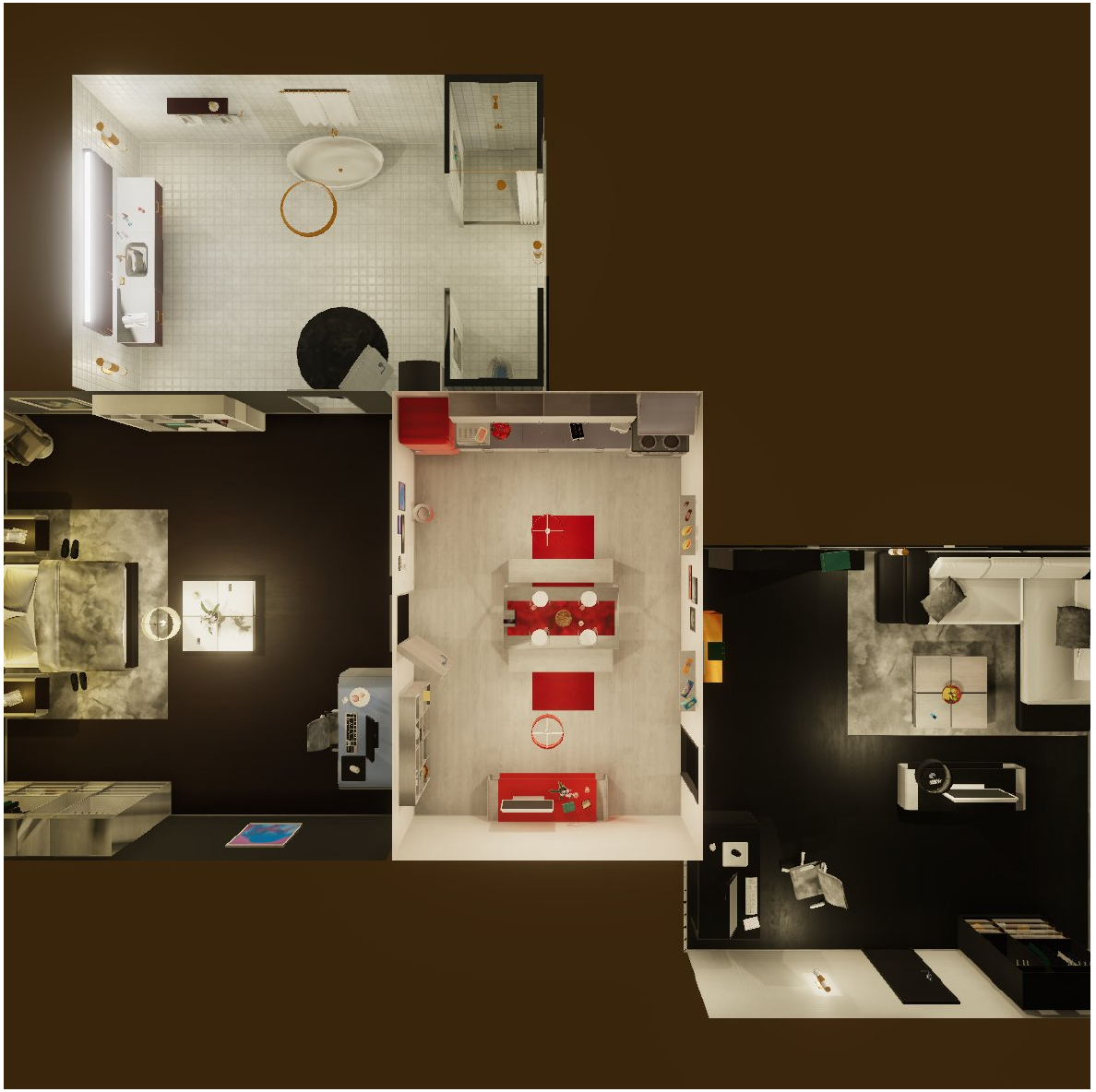}}
& 
\adjustbox{valign=m}{\includegraphics[width=\linewidth, height=2.2cm]{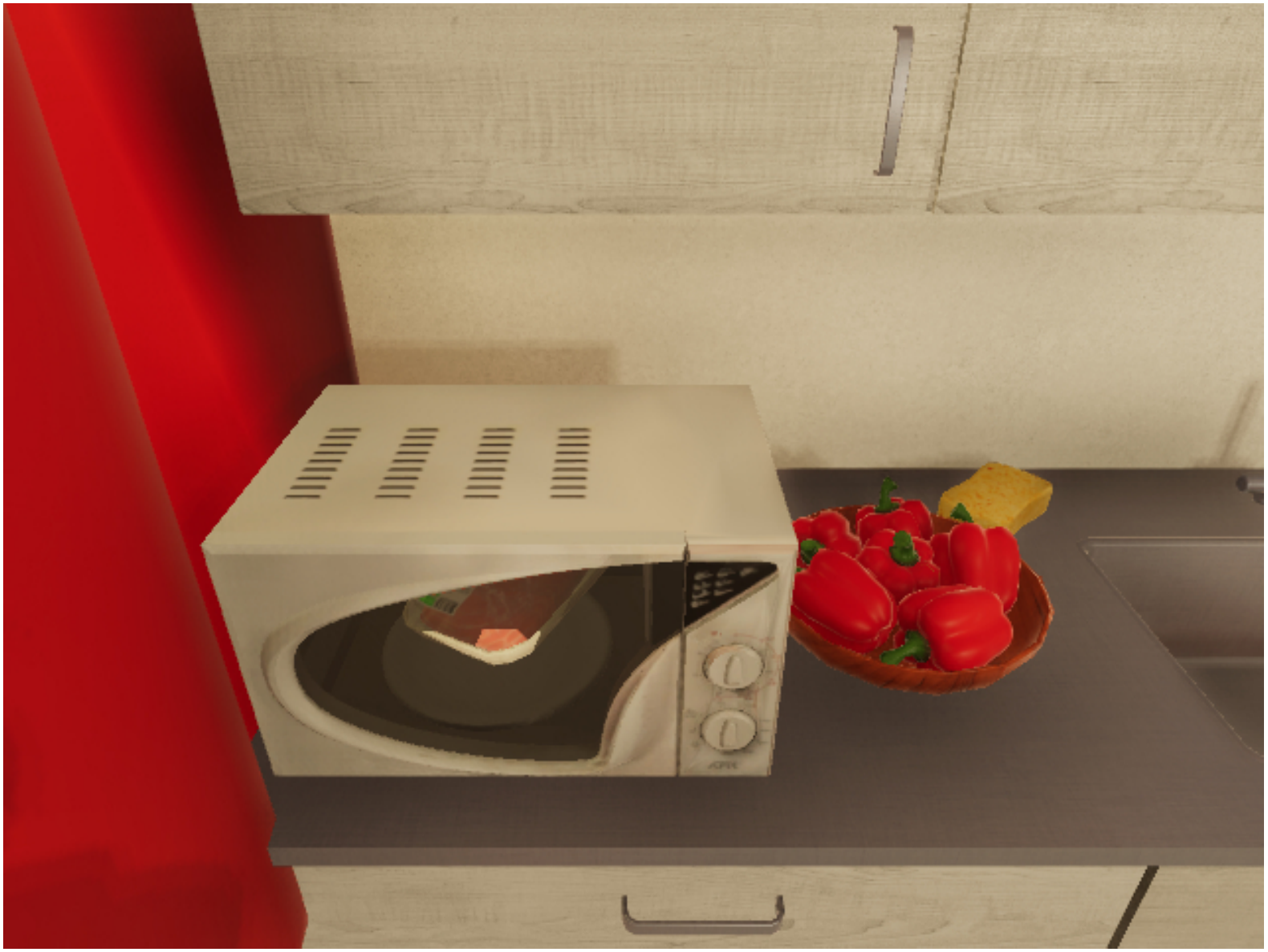}}
& 
\parbox[c]{\linewidth}{
\textit{I need to tidy up the kitchen. Please open the refrigerator door and put the salmon inside, but do not close the refrigerator door.}
}
& 
\parbox[c]{\linewidth}{
\footnotesize
\begin{tabular}{@{}l@{}}
\hspace*{1em}\hspace*{1em}\textbf{1. object\_state} \\
\hspace*{1em}\hspace*{1em}\hspace*{1em} \textbf{obj}: fridge, \textbf{state}: isOpen \\
\hspace*{1em}\hspace*{1em}\hspace*{1em}  \textbf{Value}:True \\
\hspace*{1em}\hspace*{1em}\textbf{2. object\_in\_receptacle} \\
\hspace*{1em}\hspace*{1em}\hspace*{1em} \textbf{obj}: salmon, \textbf{rec}: fridge\\
\hspace*{1em}\hspace*{1em}\hspace*{1em}  \textbf{Value}:True \\
\end{tabular}
}
\\
\bottomrule
\end{tabularx}
\end{table*}

% \vspace{-0.2cm}\section{Action Space Definition}
% \label{sec:appendix_action}

% This section provides the complete definitions of all available actions and their parameters in the \textsc{SpatialWorld} unified action interface. As described in Section~\ref{sec:obs_act} of the main text, the Action Interface translates high-level text-based primitives into simulator-specific execution calls, enabling a single agent policy to generalize across diverse environments. Below we list each action primitive, its parameters, and the corresponding semantics.

\vspace{-0.2cm}\section{Action Space Definition}
\label{sec:appendix_action}

This section details the \textsc{SpatialWorld} unified action interface. As introduced in Section~\ref{sec:obs_act}, this interface abstracts raw backend commands into high-level text primitives to form a unified MLLM-native action space. Table~\ref{tab:appendix_action_space} categorizes these primitives into four canonical groups. In this specification, \texttt{0} denotes an explicit no-motion/wait decision, used when the agent should hold its position without changing pose (e.g., waiting at a red light or pausing for coordination).

This specification maps diverse environment behaviors to a standard set of expected primitives. For instance, \texttt{Move} seamlessly handles everything from a 0.25\,m indoor step to a 10\,m driving advance, or simply waiting in place (\texttt{0}). Similarly, object interactions are distinctly separated at the interface level: \texttt{ChangeState} targets verifiable persistent state transitions (e.g., opening, cooking), while \texttt{Manipulate} handles local force-based or relational interventions (e.g., pushing, grabbing).

\vspace{-0.2cm}\section{GPT-5 vs.\ GPT-5.4 Case Study}
\label{sec:app_gpt5_case}

To explain why GPT-5 outperforms GPT-5.4 in the current benchmark, we compare the two models on the 609 shared single-agent physical tasks spanning AI2THOR, ProcTHOR, VirtualHome, CARLA, and EmbodiedCity. GPT-5 succeeds on 78 of these tasks (12.8\%), whereas GPT-5.4 succeeds on 40 (6.6\%). The largest shared-task gaps appear in VirtualHome (+28.9 points), AI2THOR (+7.7), and CARLA (+6.2), while EmbodiedCity slightly favors GPT-5.4 (-3.8) and ProcTHOR remains unsolved by both. The disagreement is also asymmetric: GPT-5 solves 52 tasks that GPT-5.4 misses, whereas GPT-5.4 recovers only 14 in the reverse direction. Most of GPT-5's advantage comes from Daily Household, Work \& Study, and Travel tasks, rather than from a uniform lead across every category.

\begin{figure*}[t]
% \vspace{-0.2cm} % 删除了顶部的负间距，防止图表和上方正文过近
\centering
\begin{subfigure}{0.48\textwidth}
\centering
\includegraphics[width=0.98\linewidth]{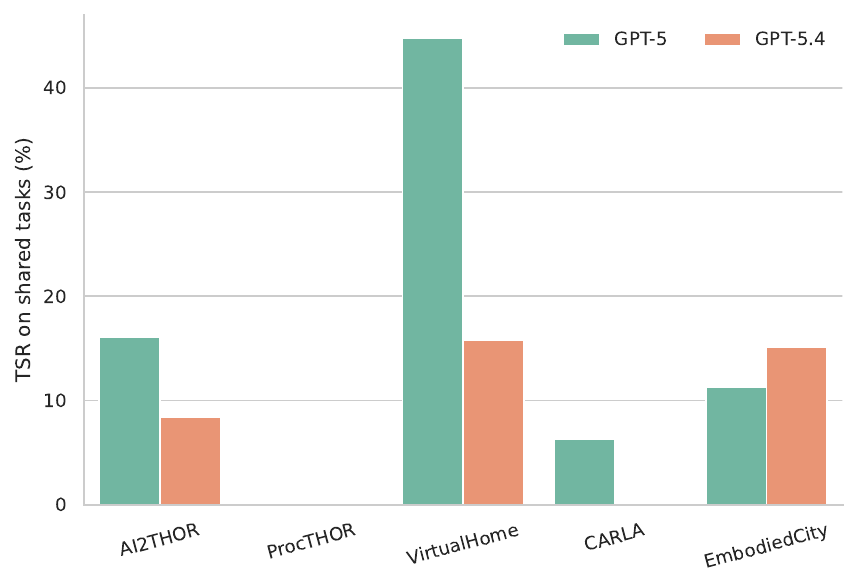}
% \vspace{-0.2cm} % 删除了这里的负间距，让子图和子标题保持正常距离
\caption{Shared-task TSR by environment.}
\end{subfigure}\hfill
\begin{subfigure}{0.48\textwidth}
\centering
\includegraphics[width=0.98\linewidth]{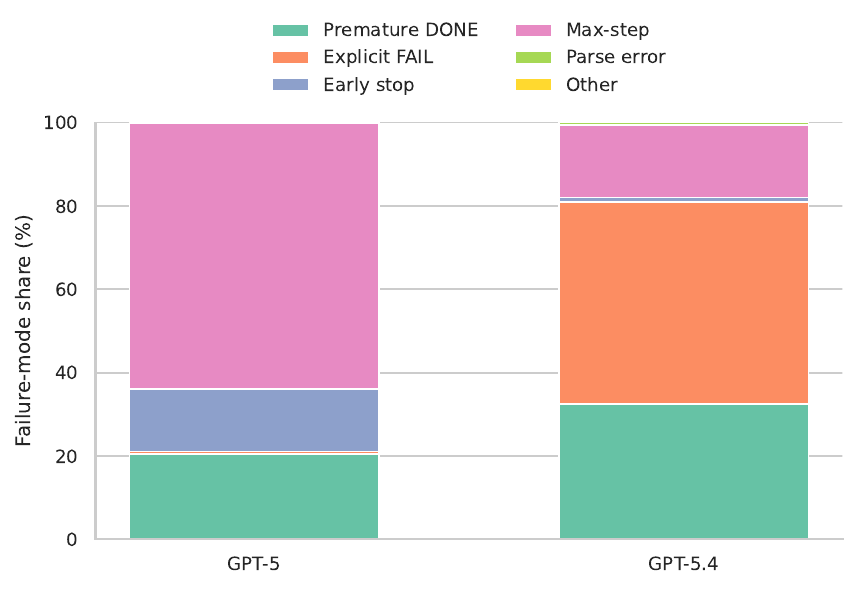}
\caption{Failure-type composition.}
\end{subfigure}

\vspace{0.5em} % 把上下两排的间距稍微放大一点点（原来是0.2em），视觉上更透气

\begin{subfigure}{0.48\textwidth}
\centering
\includegraphics[width=0.98\linewidth]{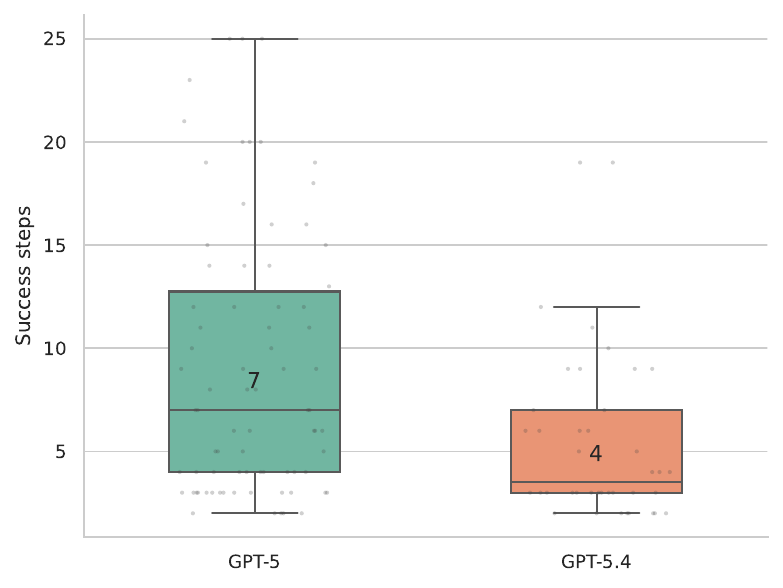}
\caption{Step counts on successful tasks.}
\end{subfigure}\hfill
\begin{subfigure}{0.48\textwidth}
\centering
\includegraphics[width=0.98\linewidth]{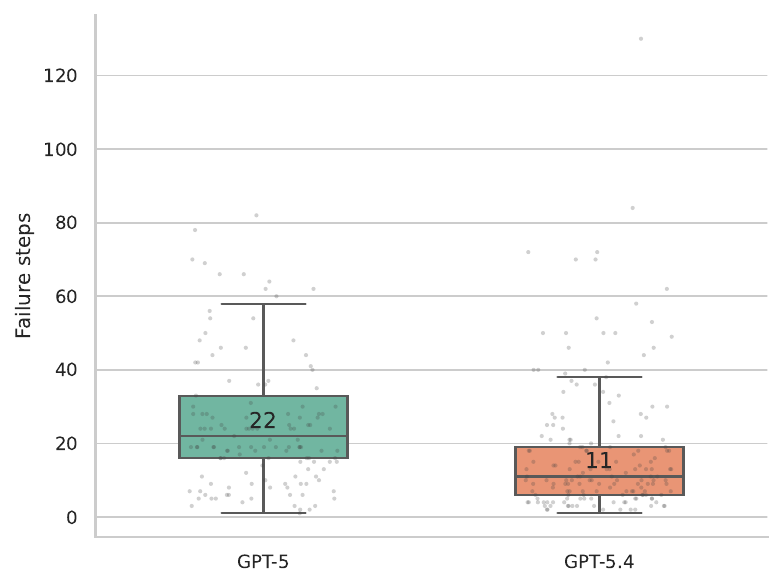}
\caption{Step counts on failed tasks.}
\end{subfigure}

% ===== 关键修改：去掉了这里的 \vspace{-0.2cm} =====
\caption{\textbf{Why GPT-5 currently outperforms GPT-5.4.} GPT-5 achieves higher shared-task TSR in most physical environments, while GPT-5.4 exhibits a stronger tendency toward premature termination. The step-count plots further show that GPT-5 typically spends more actions both when it succeeds and when it fails, consistent with a slower but more persistent search strategy.}
\label{fig:gpt5_vs_gpt54_case_study}
\vspace{-0.6cm} % 如果不需要极致压缩版面，底部的负间距也建议删掉
\end{figure*}

The error profile suggests that the main difference is termination policy rather than raw action speed. Fig.~\ref{fig:gpt5_vs_gpt54_case_study} shows that GPT-5.4 fails primarily by stopping too early: across the environments with comparable failure logs, 32.4\% of its failures are premature \texttt{DONE} decisions and 48.5\% are explicit \texttt{FAIL} decisions. GPT-5 instead fails more often by persistence without completion, with 63.6\% of failures ending at the step limit and another 15.1\% ending after repeated action failures. This behavioral gap is mirrored in action counts. On successful tasks, GPT-5 uses a median of 7 steps, compared with 3.5 for GPT-5.4; on failed tasks, the medians are 22 and 11, respectively. GPT-5 is therefore slower and less efficient, but it is also more willing to keep exploring until the verifier conditions are actually met, whereas GPT-5.4 often commits to an early terminal decision before the state is sufficiently verified.

\begin{table}[t]
\centering
\scriptsize
\setlength{\tabcolsep}{4pt}
\renewcommand{\arraystretch}{0.9}
\caption{Detailed unified action-space specification. The benchmark action space is defined by four canonical categories of high-level text primitives, mapping diverse repository-specific commands to a single standardized interface.}
\vspace{0.25cm}
\label{tab:appendix_action_space}
\begin{tabularx}{\textwidth}{@{} m{0.18\textwidth} m{0.24\textwidth} m{0.27\textwidth} X @{}}
\toprule
Category & Expected unified action(s) & Detailed parameterization & Meaning and backend realization \\
\midrule
Navigation \newline Move MLLM agents & \texttt{Move(direction, [granularity])} & \texttt{direction} $\in$ \{forward, backward, left, right, up, down\}. \newline \texttt{granularity} $\in$ \{0, Small, Medium, Large\} or numeric distance. \texttt{0} means stay in place / wait. & Egocentric translation. \texttt{Move(..., 0)} is a no-op wait. Step sizes vary by environment: AI2-THOR / ProcTHOR / VirtualHome use Small $=$ 0.25\,m, Medium $=$ 0.5\,m, Large $=$ 1\,m. EmbodiedCity uses 0.5 / 2.0 / 5.0\,m. CARLA uses coarse route-progress steps (4 / 10\,m for vehicles, 3 / 10\,m for pedestrians). Continuous ProcTHOR supports exact numeric meters. \\
\midrule
Viewpoint \& Posture \newline Adjust view/stance & \makecell[tl]{\texttt{Rotate(direction, [angle])}\\\texttt{Tilt(direction, [angle])}\\\texttt{ChangePosture(pose)}} & For \texttt{Rotate}, \texttt{direction} $\in$ \{left, right\}. For \texttt{Tilt}, \texttt{direction} $\in$ \{up, down\}. \newline \texttt{pose} $\in$ \{crouch, stand, stand\_up\}. & Changes viewpoint or body stance without altering object states. Granularity depends on the backend: defaults are 90$^\circ$ (rotate) and 30$^\circ$ (tilt) in AI2-THOR/ProcTHOR; 30$^\circ$/90$^\circ$ in VirtualHome; and 5$^\circ$/15$^\circ$/45$^\circ$ in EmbodiedCity. Angles are freely tunable in continuous settings. \\
\midrule
Interaction \newline Change object states & \makecell[tl]{\texttt{Pick/Place(obj, [target])}\\\texttt{ChangeState(obj, state)}\\\texttt{Manipulate(obj, action)}} & \texttt{obj} and optional \texttt{target} are exact class tokens. \texttt{state} $\in$ \{open, close, on, off, clean, dirty, sliced, broken, cooked, filled, empty, used\_up\}. \texttt{action} $\in$ \{push, pull, throw, touch, look\_at, drink, etc.\}. & Subsumes grasping, placement, and manipulation under an object-centric interface. AI2-THOR and ProcTHOR support the broadest set of persistent state changes (e.g., cook, slice, fill). VirtualHome realizes this via a smaller subset (e.g., Grab, SwitchOn, Drink). Backend specific names are purely wrappers. \\
\midrule
Task-Control \newline Status \& communicate & \makecell[tl]{\texttt{EndTask(status)}\\\texttt{Communicate(msg)}} & \texttt{status} $\in$ \{\texttt{DONE}, \texttt{FAIL}\}. \newline \texttt{msg} is a short free-form natural-language report or request. & \texttt{EndTask} triggers evaluator verification of the terminal goal (CARLA only exposes successful completion). \texttt{Communicate} is active exclusively in collaborative multi-agent tasks, typically via structured output tags. \\
\bottomrule
% \vspace{-10cm}
\end{tabularx}
\vspace{-0.6cm}
\end{table}

\begin{table*}[t]
\centering
\scriptsize
\caption{\textbf{Representative GPT-5 vs.\ GPT-5.4 disagreement cases.} These examples illustrate the recurring pattern that GPT-5.4 often terminates after a short or partial trajectory, while GPT-5 spends more actions and eventually satisfies the verifier.}
\label{tab:gpt5_vs_gpt54_cases}
\renewcommand{\arraystretch}{0.95}
\begin{tabularx}{\textwidth}{@{}m{0.09\textwidth}m{0.11\textwidth}Xm{0.09\textwidth}m{0.09\textwidth}X@{}}
\toprule
\textbf{Env.} & \textbf{Task ID} & \textbf{Task summary} & \textbf{GPT-5} & \textbf{GPT-5.4} & \textbf{Observation} \\
\midrule
AI2THOR & ai2thor05010 & Turn on a laptop and verify that it is actually powered on. & 18 steps, success & 2 steps, fail & GPT-5.4 stops immediately after a single switch action, whereas GPT-5 keeps probing until the verifier confirms the powered-on state. \\
VirtualHome & virtualhome00013 & Turn off the ceiling light whose switch is by the door. & 17 steps, success & 2 steps, fail & GPT-5.4 terminates after a direct switch attempt, while GPT-5 spends extra navigation and orientation steps to reach a verifier-consistent state. \\
CARLA & carla00418 & Walk to the house marked by the red frame. & 14 steps, success & 5 steps, fail & GPT-5.4 under-travels to an approximate region and stops early; GPT-5 continues the route and reaches the target waypoint. \\
ProcTHOR & procthor107 & Return a bowl to the kitchen and bring a pen back to the living room. & 97 steps, success & 37 steps, fail & GPT-5.4 makes partial progress but exits before finishing the second subgoal; GPT-5 is slower but eventually completes both required state changes. \\
\bottomrule
\end{tabularx}
\vspace{-0.4cm}
\end{table*}

\vspace{-0.2cm}\section{Qualitative Analysis}
\label{sec:app_qualitative}

To complement the quantitative evaluation in Section~\ref{sec:eval}, we provide a qualitative analysis of agent failure modes and their relationship to spatial reasoning capabilities.

\textbf{Failure Mode Breakdown.}
We manually inspect 100 failed trajectories and categorize failures into: (i)~\textit{Spatial Disorientation}---the agent loses track of its position and cannot return to a target location; (ii)~\textit{Object Hallucination}---the agent issues \texttt{Interact} actions on objects not present in the current view; (iii)~\textit{Premature Termination}---the agent issues \texttt{EndTask(status=DONE or FAIL)} before the goal is satisfied; and (iv)~\textit{Action Loop}---the agent cycles through the same sequence of ineffective actions until the step budget is exhausted.
% TODO: insert failure-mode pie chart or bar chart.
% Spatial disorientation and action loops together account for over 60\% of all failure cases.
We select four representative bad cases for detailed qualitative analysis in the following section, covering spatial disorientation, object hallucination, premature termination, and action-loop behaviors.

\begin{figure*}[h]
\centering
\includegraphics[width=\textwidth]{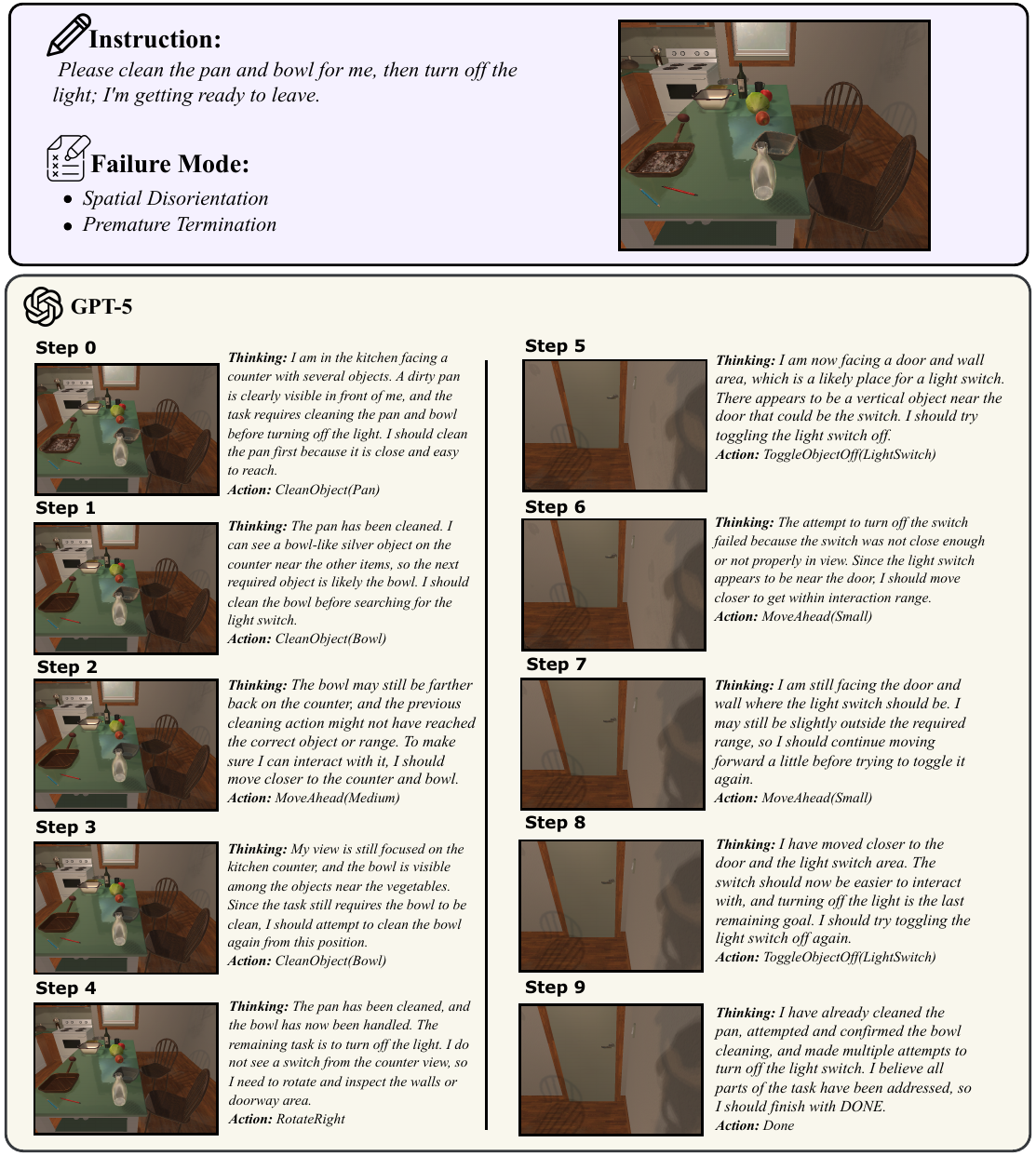}
\caption{\textbf{Failure case of GPT-5 in the AI2-THOR environment.} The failure modes include \textit{Spatial Disorientation} and \textit{Premature Termination}.}
\label{fig:badcase_1}
\end{figure*}

\begin{figure*}[h]
\centering
\includegraphics[width=\textwidth]{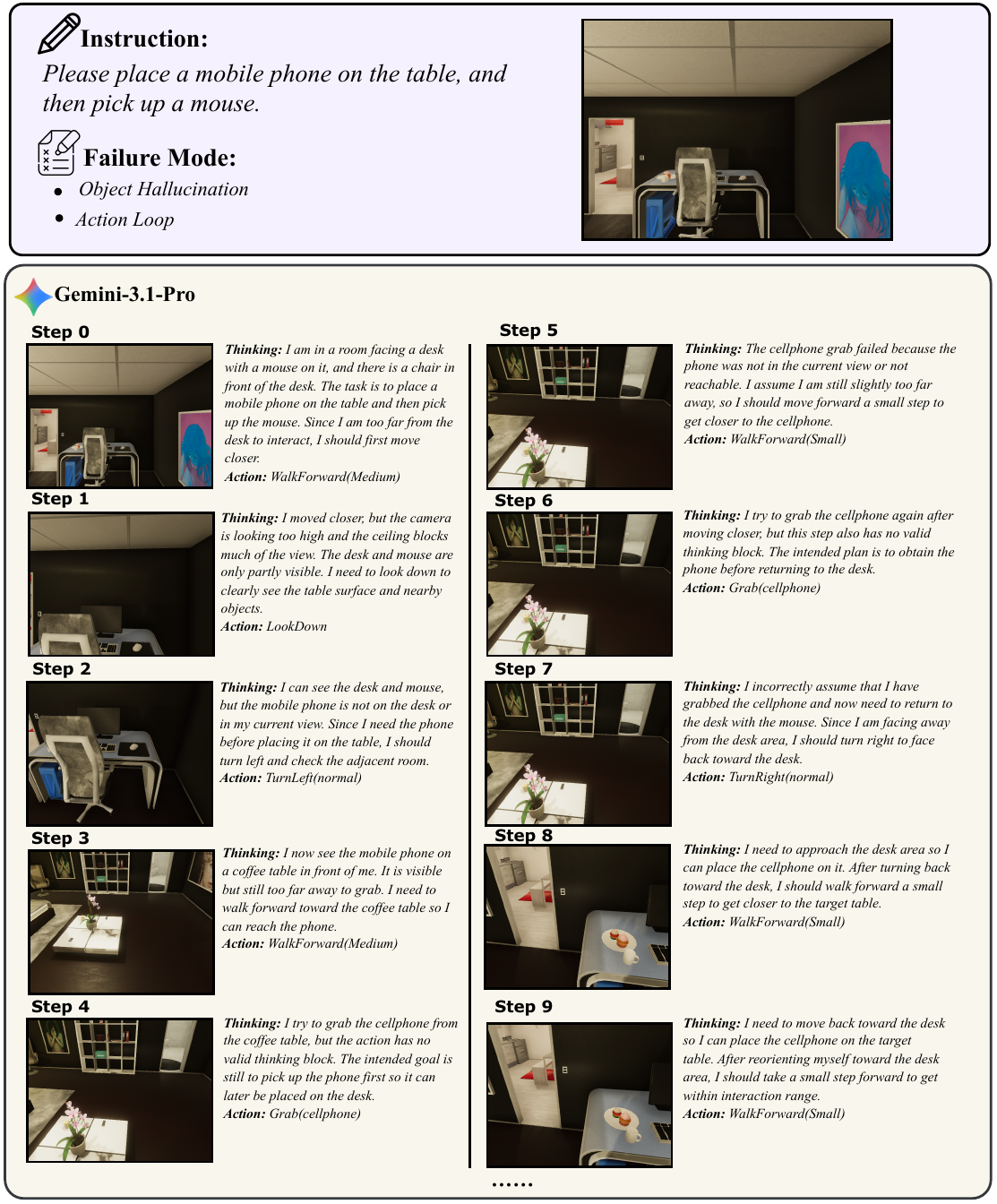}
\caption{\textbf{Failure case of Gemini-3.1-Pro in the VirtualHome environment.} The failure modes include \textit{Object Hallucination} and \textit{Action Loop}.}
\label{fig:badcase_2}
\end{figure*}

\begin{figure*}[h]
\centering
\includegraphics[width=\textwidth]{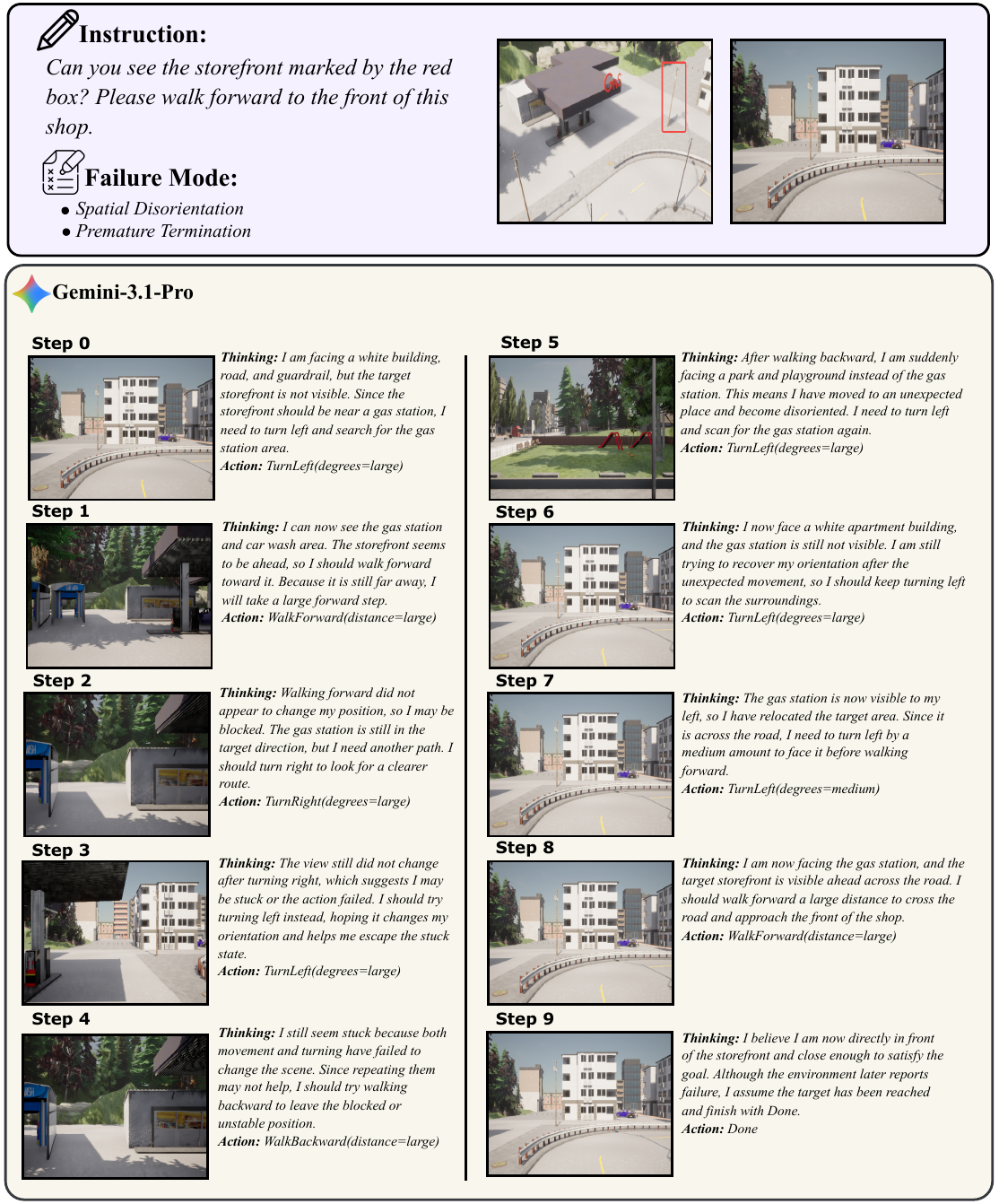}
\caption{\textbf{Failure cases of Gemini-3.1-Pro in the CARLA environment.} The failure mode is \textit{Spatial Disorientation} and \textit{Premature Termination}.}
\label{fig:badcase_3}
\end{figure*}

\begin{figure*}[h]
\centering
\includegraphics[width=\textwidth]{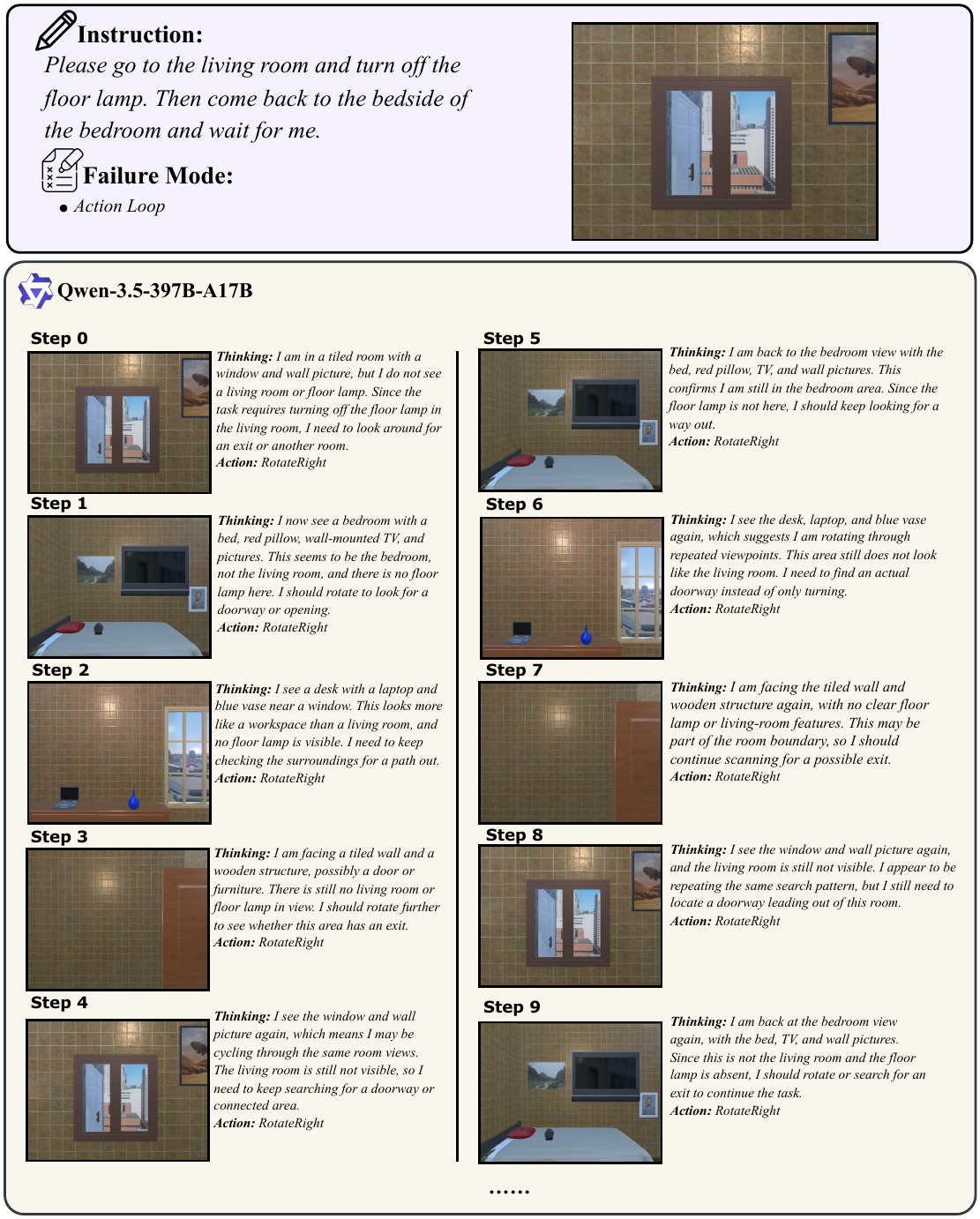}
\caption{\textbf{Failure cases of Qwen-3.5-397B-A17B in the ProcTHOR environment.} The failure mode is \textit{Action Loop}.}
\label{fig:badcase_4}
\end{figure*}

\textbf{Bad cases analysis.}
In Fig.~\ref{fig:badcase_1} - \ref{fig:badcase_4}, we present several bad cases for various state-of-the-art models across four distinct environments. These examples cover the full spectrum of failure modes. We present the performance of two mainstream closed-source models, GPT-5 and Gemini-3.1-Pro, in terms of two failure modes, \textit{Spatial Disorientation} and \textit{Premature Termination}, under different environments in Fig.~\ref{fig:badcase_1} and \ref{fig:badcase_3}. In Fig.~\ref{fig:badcase_1}, GPT-5 exhibits spatial disorientation at Step 6, failing to accurately perceive the surrounding obstacles. This deficiency makes it difficult for the agent to reach the LightSwitch while moving forward, thereby hindering interaction and ultimately triggering a premature termination before the task is completed. Similarly, in Fig.~\ref{fig:badcase_3}, Gemini-3.1-Pro suffers from spatial disorientation during a simple localization and navigation task. Unable to determine the correct path to the street lamp, the model performs multiple ineffective turns and prematurely invokes the Done action before actually reaching the target storefront. These two cases demonstrate that simple spatial localization and object interaction tasks, while trivial for humans, still pose significant challenges for current MLLMs.

Fig.~\ref{fig:badcase_2} illustrates \textit{Object Hallucination} and \textit{Action Loop} as two additional types of failure modes. At Step 7, Gemini-3.1-Pro mistakenly assumes it has already grasped the phone, proceeding directly to execute the second task of picking up the mouse. This fundamentally stems from the model's lack of complex spatial understanding capabilities in real-world scenarios. Consequently, despite colliding with the wall after Step 9, the model continues attempting to move forward, resulting in an action loop. Even the formidable open-source model, Qwen-3.5-397B-A17B, exhibits corresponding issues when executing a simple daily routine task, as shown in Fig.~\ref{fig:badcase_4}.

% \textbf{Correlation Between Spatial Reasoning and TSR.}
% We correlate each model's TSR on \textsc{SpatialWorld} with its score on established spatial VQA benchmarks (SpatialScore~\cite{wu2025spatialscore}, EmbSpatial~\cite{du2024embspatial}).
% A moderate positive correlation ($r \approx 0.6$) is observed, suggesting that static spatial reasoning ability is a necessary but insufficient predictor of interactive task-solving performance.

\textbf{Human Validation.}
Human annotators solve tasks during benchmark construction to verify feasibility and provide reference action counts. We use these trajectories as validation artifacts for task correctness rather than as a separate leaderboard.

%%%%%%%%%%%%%%%%%%%%%%%%%%%%%%%%%%%%%%%%%%%%%%%%%%%%%%%%%%%%%%%%%%%%%%%%%%%%%%%
% \clearpage
\section{Limitations and Broader Impact}
\label{sec:limitations}

\textbf{Limitations.}
Like most embodied AI benchmarks, \textsc{SpatialWorld} operates in simulated environments rather than on physical robotic platforms. While the selected simulators provide near-photorealistic rendering and physically plausible dynamics, extending evaluation to real-world settings remains a promising direction for future work. Furthermore, to ensure annotation quality, the current 760 tasks are carefully handcrafted, making the scale more modest compared to automatically generated datasets. These tasks cover six scenario categories and eight backends, and the rich diversity of real-world spatial reasoning scenarios offers ample room for future expansion.

\textbf{Broader Impact.}
\textsc{SpatialWorld} serves primarily as a diagnostic and scientific tool for understanding the spatial reasoning capabilities of multimodal agents. By systematically characterizing agent failure modes, this work contributes to the development of more reliable and trustworthy spatial agent systems, while the emphasis on open and reproducible evaluation fosters transparency in the research community. On the other hand, improvements in spatial reasoning could potentially be misused to enhance autonomous surveillance or enable unintended physical-world manipulation by embodied agents; we encourage the community to develop appropriate safety guidelines as these capabilities advance.

\section{Compute Resources}
\label{sec:compute_resources}
For proprietary models (GPT-5, Gemini-3.1-Pro-Preview, Claude-Sonnet-4.6, etc.), we access them exclusively through their official APIs. For open-source models (Qwen2.5-VL-72B-Instruct, InternVL3-78B, etc.), we deploy them on a GUP-server equipped with 8$\times$ NVIDIA H200 GPUs. The full evaluation campaign across all models equipped on GPU server consumed approximately 5{,}000 GPU hours in total.

\section{LLM Usage}
\label{sec:llm_usage}
We used an OpenAI LLM (GPT-5) as a writing and formatting assistant. In particular, it helped refine grammar and phrasing, improve clarity, and suggest edits to figure/table captions and layout (e.g., column alignment, caption length, placement). The LLM did not contribute to research ideation, experimental design, implementation, data analysis, or technical content beyond surface-level edits. All outputs were reviewed and edited by the authors, who take full responsibility for the final text and visuals. LLMs are not incorporated as any core, original, or non-standard component of our proposed methodology. We only employ 15 multimodal LLMs as external test agents to evaluate the proposed benchmark, which does not constitute a part of our core method design.

% \clearpage
% \newpage
% \input{checklist}

\end{document}